\documentclass[11pt]{article}
\PassOptionsToPackage{table}{xcolor}

\usepackage[final]{acl}

\usepackage{times}
\usepackage{latexsym}
\usepackage{comment}
\usepackage{capt-of}

\usepackage[T1]{fontenc}

\usepackage[utf8]{inputenc}

\usepackage{microtype}

\usepackage{inconsolata}

\usepackage{graphicx}
\usepackage{amsmath}

\usepackage{booktabs}
\usepackage{multirow}
\usepackage{array}
\usepackage{tabularx}

\usepackage[most]{tcolorbox}
\tcbuselibrary{skins, breakable, listings}

\newcommand{\smallra}[1][1.0]{\renewcommand{\arraystretch}{#1}}

\definecolor{maincolor}{RGB}{50, 70, 90}     
\definecolor{accentbg}{RGB}{245, 247, 250}   
\definecolor{tagbg}{RGB}{230, 235, 240}      
\definecolor{tagtext}{RGB}{40, 50, 60} 
\definecolor{linecolor}{gray}{0.85} 

\newtcbox{\tagbox}[1][]{on line,
  arc=2pt, outer arc=2pt,
  colback=tagbg, colframe=tagbg,
  colupper=tagtext, fontupper=\bfseries\scriptsize,
  boxrule=0pt, bottom=1pt, top=1pt, left=4pt, right=4pt,
  valign=center, #1}

\newcommand{\inlineitem}[2]{%
    \noindent\textbf{\color{maincolor}#1:} #2\par\vspace{3pt}%
}

\newtcolorbox{prettybox}[2][]{%
    enhanced,
    colback=white,
    colframe=maincolor,
    fonttitle=\bfseries\large,
    title={#2},
    attach boxed title to top left={yshift=-12pt, xshift=15pt},
    boxed title style={colback=maincolor, frame hidden, arc=3pt},
    boxrule=1.2pt,
    top=16pt,      
    bottom=8pt,    
    arc=3pt,
    #1
}
\title{KnowMe-Bench: Benchmarking Person Understanding for Lifelong Digital Companions}

\author{
\textbf{Tingyu Wu\textsuperscript{1,2*}},
\textbf{Zhisheng Chen\textsuperscript{1,2*}},
\textbf{Ziyan Weng\textsuperscript{4*}},
\\
\textbf{Shuhe Wang\textsuperscript{7}},
\textbf{Shuo Zhang\textsuperscript{2}},
\textbf{Sen Hu\textsuperscript{2,3}},
\textbf{Silin Wu\textsuperscript{5}},
\\
\textbf{Qizhen Lan\textsuperscript{2,6$\dagger$}},
\textbf{Huacan Wang\textsuperscript{1,2$\dagger$}},
\textbf{Ronghao Chen\textsuperscript{2,8$\dagger$}}
\\
\\
 \textsuperscript{1}UCAS,
 \textsuperscript{2}QuantaAlpha,
 \textsuperscript{3}PKU,
 \textsuperscript{4}CITYU-DG,
 \textsuperscript{5}HAINNU,
 \textsuperscript{6}{UTHealth},
 \textsuperscript{7}{NUS},
 \textsuperscript{8}IAPM (Guangdong)
\\
\small {
    \textbf{\textsuperscript{*}These authors contributed equally to this work.}
}
\\
 \small{
   \textbf{$\dagger$ Correspondence:} \href{mailto:Qizhen.Lan@uth.tmc.edu}{Qizhen.Lan@uth.tmc.edu},
      \href{mailto:chenronghao@alumni.pku.edu.cn}{chenronghao@alumni.pku.edu.cn},
   \href{mailto:wanghuacan17@mails.ucas.ac.cn}{wanghuacan17@mails.ucas.ac.cn}
 }
}

\newcommand{\BenchName}{KnowMe-Bench}
\newcommand{\BenchTokens}{\textasciitilde 4.7M} 

\begin{document}
\maketitle

\begin{abstract}
Existing long-horizon memory benchmarks mostly use multi-turn dialogues or synthetic user histories, which makes retrieval performance an imperfect proxy for person understanding. We present \BenchName, a publicly releasable benchmark built from long-form autobiographical narratives, where actions, context, and inner thoughts provide dense evidence for inferring stable motivations and decision principles. \BenchName~reconstructs each narrative into a flashback-aware, time-anchored stream and evaluates models with evidence-linked questions spanning factual recall, subjective state attribution, and principle-level reasoning. Across diverse narrative sources, retrieval-augmented systems mainly improve factual accuracy, while errors persist on temporally grounded explanations and higher-level inferences, highlighting the need for memory mechanisms beyond retrieval.
\end{abstract}

\section{Introduction}

A long-standing goal in Artificial Intelligence is to build \emph{lifelong digital companions} that can support users over extended horizons by maintaining coherent personalization, context awareness, and behavior consistent with users' evolving goals and values. Recent LLM-based agent frameworks increasingly aim at sustained interaction across sessions rather than isolated question answering \cite{park2023generativeagents,zhong2024memorybank,packer2023memgpt}. In this setting, the central capability is \textbf{\emph{person understanding}}: a companion should form and update an internal model of the user that supports explanation (why a choice was made), anticipation (what the user is likely to prefer next), and alignment (what the user seeks to pursue or avoid).

Importantly, \emph{memory} is a necessary substrate but not a sufficient definition of person understanding. A system can store and retrieve facts yet still fail to infer stable principles, connect distant experiences to present reactions, or explain recurring decision patterns. This paper therefore asks: \emph{how should we benchmark person understanding as an evidence-grounded inference problem over lived experience, rather than as retrieval over a fact database?}


\begin{figure*}[!t]
    \centering
    \includegraphics[width=\linewidth]{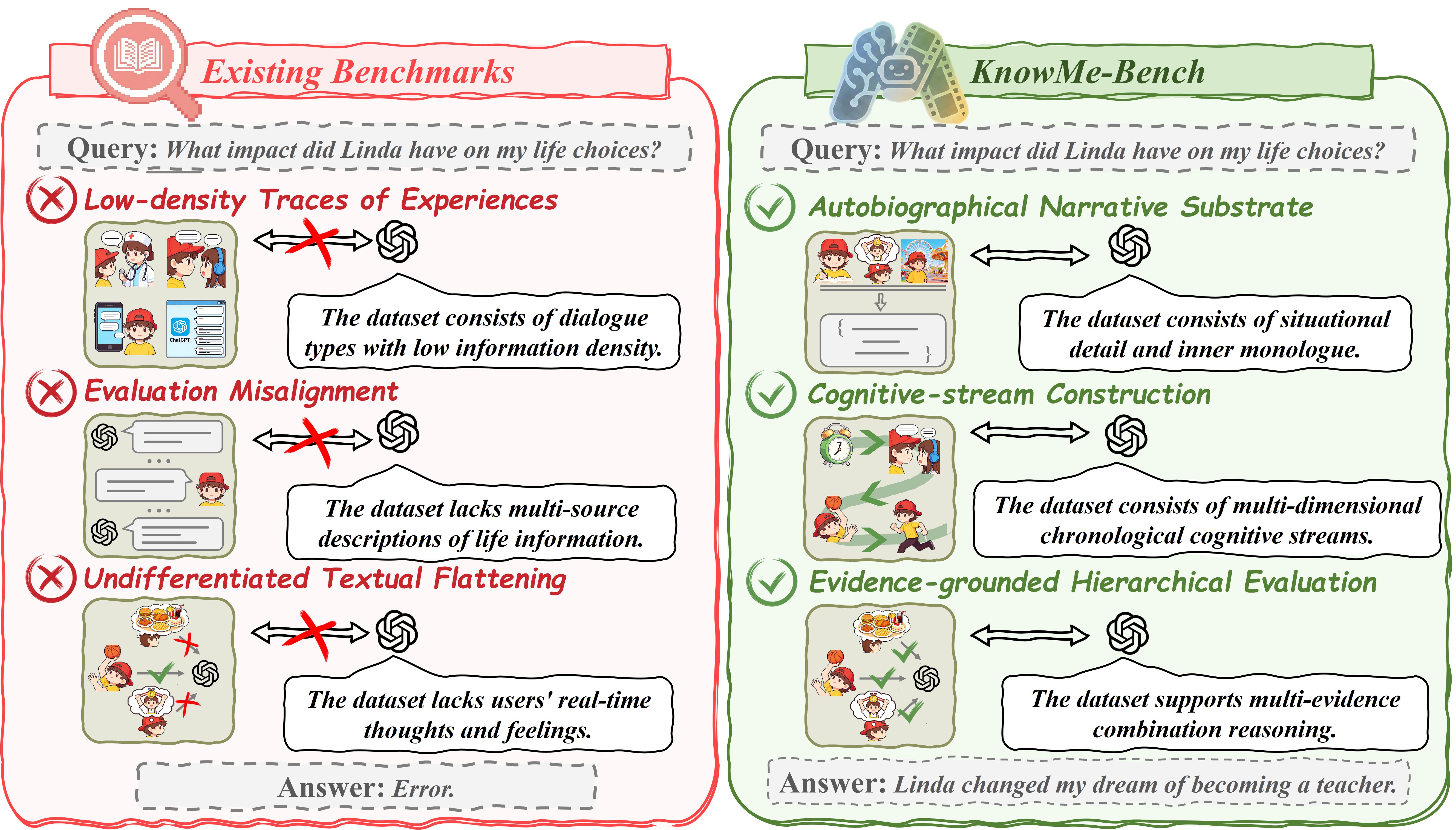}
    \caption{Comparison of benchmark substrates. Existing long-horizon memory evaluations typically rely on sparse dialogue traces or synthetic interaction histories, which provide limited support for evidence-grounded person modeling. KnowMe-Bench instead uses autobiographical narratives with aligned actions, context, and inner thoughts, enabling flashback-aware reconstruction and hierarchical evaluation from factual recall to interpretive reasoning.}
    \label{fig:difference}
\end{figure*}

Despite rapid progress on long-horizon agent evaluation, we identify two gaps that prevent existing benchmarks from directly measuring person understanding.

\paragraph{\emph{(G1) Evaluation misalignment: retrieval proxies $\neq$ person understanding.}}
Most benchmarks focus on retrieval, temporal ordering, knowledge updates, conflict handling, or longitudinal state tracking across sessions \cite{wu2025longmemeval,maharana2024locomo,hu2025memoryagentbench,castillo_bolado2024beyondprompts,tan2025membench,hu2026clonemem}. These tasks are necessary, yet they do not directly test whether an agent can infer and use an implicit \emph{person model}---e.g., motivations and avoidance goals, stable principles, evolving self-concepts, relationship structure, and affective triggers---to explain or anticipate behavior. In addition, ``deep'' questions without evidence constraints invite free-form speculation.

\paragraph{\emph{(G2) Data Substrate Misalignment: Low-Density and Decontextualized Experience Traces}}
Most scalable benchmarks construct user histories from chat logs, synthetic events, or model-generated interactions
\cite{maharana2024locomo,castillo_bolado2024beyondprompts,wu2025longmemeval}.
Although efficient, such substrates fail to support person understanding due to two structural limitations.
\textbf{\emph{(G2a) Density loss}}: experiences are compressed into sparse traces, weakening the coupling between observable actions and the internal deliberation that gives them personal significance.
\textbf{\emph{(G2b) Structure loss}}: heterogeneous experiential signals are flattened into undifferentiated text, erasing modality cues and temporal alignment needed for long-horizon attribution.
Accordingly, a benchmark for person-model inference must approximate the \emph{multimodal} organization of lived experience; when represented textually, this entails explicit separation of distinct \emph{textual modalities} rather than a single narrative surface.
This view aligns with autobiographical memory and narrative identity theories, which emphasize that stable self-knowledge emerges from temporally structured, subjectively interpreted experience
\cite{conway2000sms,mcadams2013narrativeidentity}.

To bridge these gaps, we introduce \textbf{KnowMe-Bench}, a benchmark for evaluating \emph{evidence-grounded person-model inference} from long-form autobiographical experience.
We operationalize this goal through three design modules (M1--M3), each explicitly aligned with the identified gaps.

\paragraph{\emph{(M1) Autobiographical narrative substrate (addresses G2a).}}
KnowMe-Bench uses autobiographical narratives that retain the joint expression of external events and internal interpretation, yielding high-density evidence suitable for person-model inference
\cite{conway2000sms,mcadams2013narrativeidentity}.

\paragraph{\emph{(M2) Cognitive-stream reconstruction with mnestic realignment (addresses G2b; supports G2a).}}
To make evidence usable for long-horizon attribution, we reconstruct narratives into a chronological cognitive stream anchored by explicit
timestamps and locations. We decompose the text into five fields: (1) visual observations, (2) auditory inputs, (3) situational context,
(4) accessible background knowledge, and (5) inner monologue. This representation improves evidence granularity (supporting G2a) and enables
\emph{mnestic realignment}: present-time mnemonic triggers remain anchored in the current timeline, while recalled content is relocated to its
chronological origin, restoring temporal and causal structure (addressing G2b).

\paragraph{\emph{(M3) Evidence-grounded hierarchical evaluation with expert verification (addresses G1; leverages M2).}}
To directly measure person understanding, we propose a three-tier evaluation suite:
\emph{Tier 1: Factual extraction}, \emph{Tier 2: Subjective state attribution}, and \emph{Tier 3: Decision and principle reasoning}.
Tiers 2--3 require (i) a concise inference and (ii) an explicit evidence set of supporting events in the aligned timeline, ensuring auditability and discouraging free-form speculation. Deep items are produced and cross-validated by trained annotators against the
reference aligned timeline.

\paragraph{Baselines and diagnostics.}
We provide baselines spanning long-context prompting, retrieval-augmented agents, and external memory-store / agentic-memory systems \cite{packer2023memgpt,zhong2024memorybank,chhikara2025mem0,xu2025amem}.
These results enable diagnostic comparison of memory mechanisms and quantify the gap between retrieval-oriented competence and person-model inference. In summary, we make three contributions:
\begin{itemize}
    \item \textbf{Benchmarking person understanding.} We formalize person understanding for lifelong digital companions as an \emph{auditable person-model inference} problem over long-horizon experience, and introduce \BenchName, a publicly releasable benchmark built from autobiographical narratives (\BenchTokens~tokens).
    \item \textbf{High-density, structured experience representation.} We construct flashback-aware, time-aligned lifelogs via cognitive-stream reconstruction with multiple textual modalities and mnestic realignment.
    \item \textbf{Hierarchical evaluation and diagnostics.} We propose an evidence-linked, three-tier evaluation protocol with expert verification and provide diagnostic baselines across representative agent designs.
\end{itemize}

\section{Related Work}

\begin{figure*}[t]
    \centering
    \includegraphics[width=\linewidth]{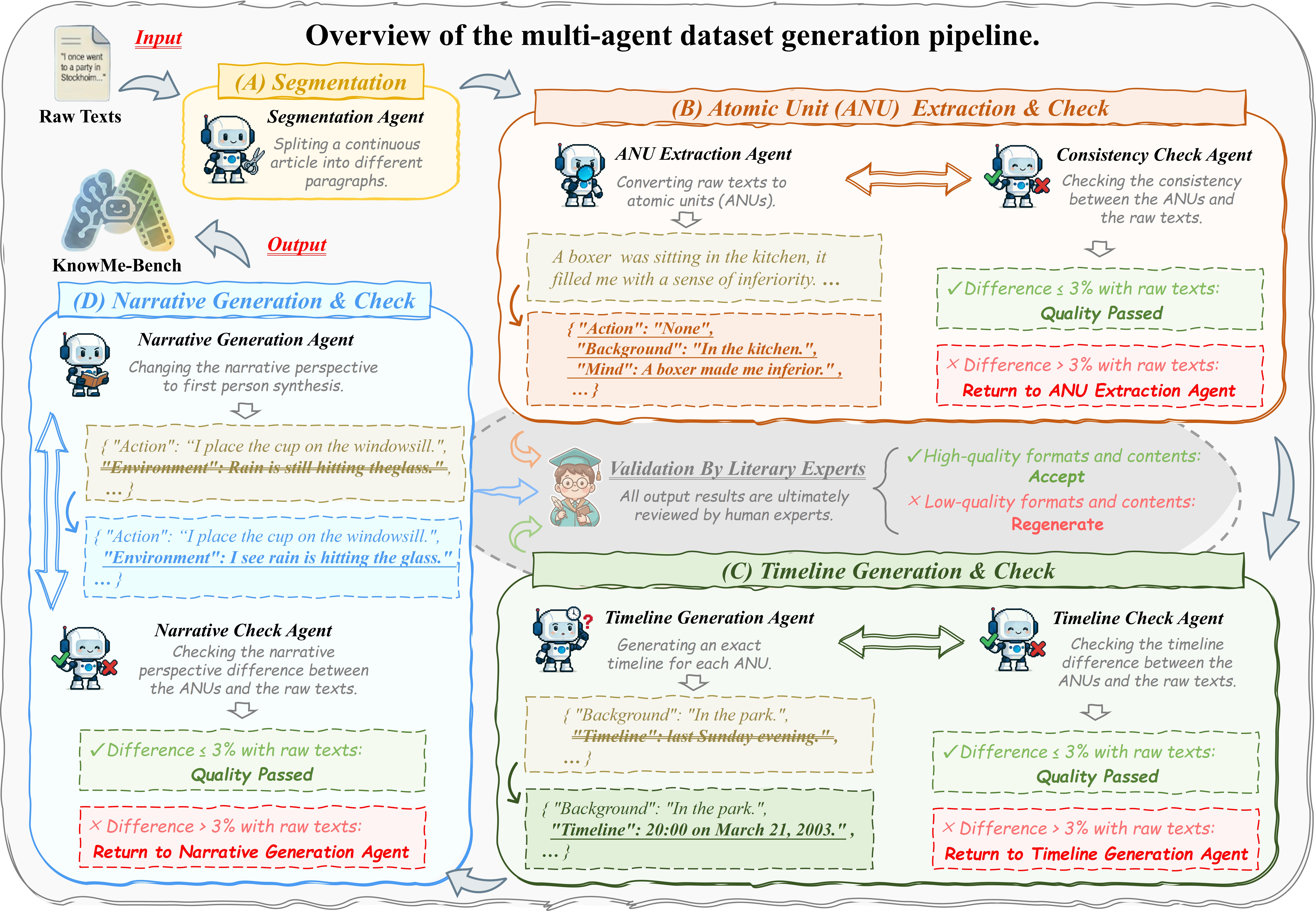}
    \caption{\textbf{Overview of the multi-agent dataset generation pipeline.} The framework transforms unstructured raw narratives into the structured KnowMe-Bench benchmark through four sequential stages: (A) Segmentation, (B) Atomic Unit (ANU) Extraction, (C) Timeline Generation, and (D) Narrative Generation. To ensure data fidelity, each generative module is paired with a specific Check Agent that enforces a ``Verify-and-Revise'' loop, culminating in final validation by human literary experts.}
    \label{fig:pipeline}
\end{figure*}

\paragraph{Evaluation of Long-Term Memory Agents.}
The evaluation of memory in LLM-based agents has evolved from effective context window tests \cite{hu2025memory} to multi-turn interactions that assess memory consolidation \cite{chhikara2025mem0, li2025memos}. Recent benchmarks focus on the agent's ability to update specific facts or track entity states over distinct conversational turns \cite{zhong2024memorybank}. More recent work also moves beyond purely conversational traces: CloneMem evaluates long-term memory over diaries, social media posts, and emails in AI-clone settings \cite{hu2026clonemem}, while concurrent work organizes personalized memory by event time rather than dialogue time, constructing semantically grounded durative memories for personalized agents \cite{su2026beyonddialoguetime}. However, these evaluations predominantly treat memory as a temporal retrieval or state-tracking problem over personal traces, prioritizing recall fidelity and longitudinal consistency over interpretative reasoning. Current benchmarks often overlook autobiographical reasoning, where an agent must infer implicit information, such as stable principles or affective triggers, from long-horizon causal chains rather than explicit statements.

\begin{table*}[!t]
    \centering
    \small
    \setlength{\tabcolsep}{5.5pt}
    \renewcommand{\arraystretch}{1.12}
    \begin{tabularx}{\textwidth}{>{\raggedright\arraybackslash}p{0.18\textwidth}>{\raggedright\arraybackslash}p{0.27\textwidth}>{\raggedright\arraybackslash}p{0.22\textwidth}>{\raggedright\arraybackslash}X}
        \toprule
        \textbf{Benchmark} & \textbf{Experience substrate} & \textbf{Temporal representation} & \textbf{Primary evaluation focus} \\
        \midrule
        LongMemEval~\cite{wu2025longmemeval} & Multi-session assistant dialogues & Dialogue order and session boundaries & Long-horizon conversational recall \\
        LoCoMo~\cite{maharana2024locomo} & Grounded long conversations & Event graphs plus dialogue sessions & Very long conversational consistency \\
        MemBench~\cite{tan2025membench} & Interactive agent scenarios & Interaction turns & Factual plus reflective memory behaviors \\
        CloneMem~\cite{hu2026clonemem} & Diaries, social posts, and emails & Longitudinal personal traces & Tracking evolving personal states for AI clones \\
        \textbf{KnowMe-Bench} & \textbf{Autobiographical narratives} & \textbf{Flashback-aware chronological stream} & \textbf{Evidence-grounded interpretive reasoning over long-horizon experience} \\
        \bottomrule
    \end{tabularx}
    \caption{Comparison with representative long-term memory evaluations. KnowMe-Bench differs most clearly in its flashback-aware chronological representation and in targeting evidence-grounded interpretive reasoning rather than only conversational recall or state tracking.}
    \label{tab:benchmark_compare}
\end{table*}

\paragraph{Benchmarks for Person Modeling and Psychology.}
Research on "Persona Agents" typically relies on static profiles or role-playing descriptions provided in the system prompt \cite{sun2025persona, kroczek2025influence, chen2025persona}. While some studies incorporate psychometric evaluations like MBTI or Big Five \cite{brickman2025large, ke2025exploring, szymanski2025limitations}, they generally use these frameworks as rigid templates to steer generation. Such static profiling fails to capture the complexity of human behavior, which is inherently context-dependent and evolves over time. Furthermore, the data substrates used in these tasks are often synthetic chat logs or simulated sandbox environments \cite{cheng2025omnichat, chou2025visionarena, nguyen2026generative}. These sources lack the sensory grounding and introspective density characteristic of complex autobiographical narratives, limiting the evaluation of deep person modeling.

\paragraph{Timeline Construction and Narrative Processing.}
Constructing structured timelines from unstructured text is critical for grounding agent memory. Traditional timeline generation (TLG) often assumes a linear progression of events or relies on simplified timestamp extraction \cite{liu2025etimeline, qorib2025just}. Such linear assumptions are insufficient for processing complex personal accounts, which frequently contain non-linear temporal structures like flashbacks and mental time travel. Naive ingestion of such narratives results in causal scrambling, where past events are incorrectly anchored to the present context \cite{fatemi2024test, maharana2024evaluating}. Unlike stochastic rewriting approaches that risk hallucination, methods that model cognitive primitives and flashback-aware alignment are necessary to preserve the temporal-causal integrity of the source material.

\section{Methodology}

\subsection{Overview}
We propose \textbf{KnowMe-Bench}, a framework designed to enable evidence-grounded person-model inference over long-horizon autobiographical experience. To address the challenges of low-density evidence and non-linear narration, we construct a \emph{flashback-aware chronological cognitive stream} from raw narratives.
As illustrated in Figure~\ref{fig:pipeline}, our pipeline operates via a four-stage multi-agent workflow (Modules A--D). We enforce a \textbf{``Faithfulness-First''} principle: non-generative stages rely on index-based extraction, while generative stages are guarded by a generic \emph{Verify-and-Revise} protocol (detailed in Appendix~\ref{app:verification}) to prevent hallucination and maintain strict adherence to the source text.

\subsection{Stage I: Context-Aware Segmentation (Module A)}
Autobiographical narratives are structurally heterogeneous. To preserve causal micro-structure, Module A functions as a deterministic \textbf{semantic boundary detector}. Instead of fixed-length chunking, it identifies natural boundaries (e.g., scene transitions) and slices the raw text by indices. This purely extractive approach ensures the verbatim preservation of the original content, providing a faithful input for downstream processing.

\subsection{Stage II: Atomic Narrative Unit (ANU) Extraction (Module B)}
To expose micro-evidence, Module B decomposes raw segments into \textbf{Atomic Narrative Units (ANU)}—the smallest auditable carriers of experience.
We formally define an ANU as a tuple:
\begin{equation}
U = ( \mathrm{id},\ t^{\text{anch}},\ \ell,\ C ),
\end{equation}
where $\mathrm{id}$ is the unique identifier, $t^{\text{anch}}$ is the temporal anchor, $\ell$ is the mandatory location, and $C$ is a structured cognitive record containing five primitives: \emph{Action, Dialogue, Environment, Background,} and \emph{Mind}.
To ensure global retrievability in million-token contexts, $t^{\text{anch}}$ is not restricted to a coarse verbatim date string. When the source text provides underspecified or scene-level time expressions, the timeline agent synthesizes micro-granularity anchors consistent with the local narrative flow (e.g., second-level offsets within a scene), so that each ANU is indexed by the globally unique tuple $(\mathrm{id}, t^{\text{anch}}, \ell, C)$.
During evaluation, models are not required to reproduce the literal timestamp token. Instead, temporal tasks score whether the model recovers the correct duration, relative order, or trigger--event linkage induced by these anchors.
To ensure granularity, we impose hard constraints on the complexity of $C$ (e.g., decomposing abstract states into observable micro-behaviors), ensuring the substrate captures the high-density ``micro-texture'' of memory.

\subsection{Stage III: Flashback-Aware Temporal Realignment (Module C)}
Standard timestamp extraction fails on narratives containing nested temporal structures (e.g., flashbacks). Module C restores causal structure via a \textbf{Mnestic Realignment Protocol}.
\begin{itemize}
    \item \textbf{Mnestic Separation:} We conceptually separate each unit into the \emph{Event Content} ($C_{\text{event}}$, to be relocated to its historical origin) and the \emph{Mnemonic Trigger} ($T_{\text{trigger}}$, to remain anchored in the present stream of consciousness).
    \item \textbf{Stack-Based Alignment:} We employ a stack-based state machine to track nested contexts. The system predicts alignment actions (e.g., \textsc{Push} for entering flashbacks, \textsc{Pop} for returning) to reorder events chronologically while preserving the narrator's psychological timeline. (State transition rules and action semantics are detailed in Appendix~\ref{app:verification-feedback}.)
\end{itemize}

\subsection{Stage IV: Narrative Instantiation and Validation (Module D)}
To produce a queryable first-person record without flattening the structure, Module D acts as an \textbf{Embodied Decoder}. It performs component-wise subjectivization, transforming objective descriptors in the ANU into immediate sensory experiences (e.g., ``Rain hits glass'' $\rightarrow$ ``I see rain hitting glass'').
Final validation is conducted by human literary experts to ensure the dataset serves as a reference-quality benchmark, routing any detected errors (e.g., emotional flattening) back to the specific module for revision.

\section{Evaluation Framework}
\label{sec:evaluation}

To comprehensively assess the agent's capabilities from factual retention to literary reasoning, we introduce the \textbf{KnowMe-Bench} evaluation suite. It consists of 7 distinct tasks hierarchically categorized into three cognitive levels.

\subsection{Level I: Precision \& Factuality (The ``Memory'' Layer)}
This level evaluates the model's ability to precisely retrieve entities and temporal details from the long-context timeline ($Q_{\tau}$). \textbf{Task 1 (Context-Aware Information Extraction)} tests complete entity recall under strict spatiotemporal constraints. \textbf{Task 2 (Adversarial Abstention)} uses ``Mismatching Trap'' queries to verify that the model refuses to answer when entities or causal links are deliberately distorted. \textbf{Task 3 (Temporal Reasoning)} probes duration estimation and the ability to recover chronological order rather than narrative presentation order in flashback-heavy passages.

\subsection{Level II: Narrative Logic \& Causality (The ``Reasoning'' Layer)}
This level requires understanding logical connections and non-linear transitions. \textbf{Task 4 (Logical Event Ordering)} asks the model to order events along non-temporal semantic dimensions such as escalation of danger or emotional intensity. \textbf{Task 5 (Mnestic Trigger Analysis)} evaluates whether the model can identify the sensory cue or associative trigger that shifts consciousness from the present scene into a recalled memory.

\subsection{Level III: Interpretive Insight (The ``Insight'' Layer)}
The most challenging tier targets subtext and long-range self-explanation. \textbf{Task 6 (Mind-Body Interaction)} asks the model to reconcile external behavior with internal state, especially in ironic or self-contradictory moments. \textbf{Task 7 (Expert-Annotated Insight)} consists of expert-curated open-ended questions about motives, identity construction, and enduring decision principles, and serves as the strongest diagnostic for deep person understanding.

\subsection{Scoring Protocol: LLM-as-a-Judge}
\label{sec:scoring-protocol}
Given the subjective nature of literary analysis, purely overlap-based metrics are insufficient. We implement a rigorous \textbf{LLM-as-a-Judge} protocol (utilizing GPT-4o) with strict rubric constraints (Scale 0-5).

\paragraph{Scoring Dimensions.} 
The evaluation rubrics are tailored to task type. For factual tasks ($T_1, T_2, T_3$), the judge scores \textbf{Entity Accuracy} and \textbf{Value Precision}, with $T_2$ granting full marks only for correct abstention. For logic tasks ($T_4, T_5$), the judge evaluates \textbf{Sequence Correctness} and the \textbf{Validity of Reasoning}, namely whether the answer identifies the correct causal trigger. For insight tasks ($T_6, T_7$), scoring focuses on the \textbf{External-to-Internal Mapping}; a full score requires capturing the specific \textbf{core metaphors} in the reference answer rather than generic emotional descriptions.

\paragraph{Metric Reliability.} 
For each task, the final score is the average of the rubric-based scores. Because annotators provide answers rather than direct scalar ratings, pairwise human--human kappa is not the most informative reliability quantity here. Instead, we measure alignment between the rubric-based judge and expert consensus on a held-out subjective subset. This judge-versus-expert agreement is substantial ($\kappa > 0.75$), indicating that the scoring pipeline tracks expert grading conventions under the same blinded protocol later used for model outputs.

\section{Experiments}
\label{sec:experiments}

To validate the effectiveness of KnowMe-Bench in distinguishing between retrieval capabilities and person understanding, we conducted extensive evaluations across representative long-horizon memory systems.

\subsection{Experimental Setup}

\paragraph{Datasets and Narrative Modalities.}
We use the full KnowMe-Bench corpus (4.7M tokens) spanning three structurally distinct narrative regimes and a total of \textbf{2,580 evaluation queries}. Dataset~1 uses Knausg\aa rd's \textit{My Struggle} (1.15M tokens) and emphasizes flashbacks and mnestic triggers; Dataset~2 uses the \textit{Neapolitan Novels} (1.76M tokens) and emphasizes linear causal tracking with high-frequency entity updates; Dataset~3 uses Proust's \textit{In Search of Lost Time} (1.30M tokens) and emphasizes introspective passages and abstract internal monologue.

\paragraph{De-identification \& Ethics.}
We apply a privacy pipeline (detailed in the appendix) to remove PII while preserving narrative structure for evaluation.

\paragraph{Resources.}
The benchmark, generation pipeline, and evaluation code are released at \href{https://github.com/QuantaAlpha/KnowMeBench}{GitHub} under MIT; the dataset mirror is hosted at \href{https://huggingface.co/datasets/realty2333/knowMe-Bench}{Hugging Face} under Apache-2.0.

\paragraph{Human Evaluation Protocol.}
We ran a 3-expert human study spanning all three evaluation levels. Annotators answered benchmark questions with access to the aligned cognitive stream and, when needed, the original source text; their responses were then scored by the same blinded LLM-as-a-Judge pipeline used for model outputs. Experts reach 96.5/88.0/83.5 on Levels~I/II/III, versus 75.4/62.5/22.6 for the best model, leaving a wide gap on the interpretive tier. Consistent with Section~\ref{sec:scoring-protocol}, the reported $\kappa > 0.75$ measures judge-versus-expert alignment on the subjective subset rather than pairwise annotator agreement. Full procedural details appear in Appendix~\ref{app:human_eval_details}.

\paragraph{Model Architecture \& Baselines.}
We distinguish between the \textit{Inference Model} (generation and reasoning) and the \textit{Embedding Model} (vector retrieval). Our inference backbones are Qwen3-32B (long-context), GPT-5-mini, DeepSeek-R1, and Gemini-3 Pro. Across these backbones we evaluate a consistent Base / Naive-RAG / Mem0 / MemOS protocol; on Qwen3-32B we additionally report A-Mem to test whether reflective memory rewriting helps deep person-level inference. Naive RAG ($k=50$) is the standard dense-retrieval baseline, Mem0 represents structured entity-state memory, and MemOS represents a log-based chronological memory architecture.

\paragraph{Base Models vs. Memory Systems.}
For plain base models, the full narrative usually exceeds the effective input window. We therefore use a truncation protocol that preserves the initial setup and the most recent context at inference time. Memory systems, by contrast, do not ingest the full narrative monolithically; they retrieve a task-conditioned top-$k$ set of fragments or chronological logs from external storage. This distinction is important for interpreting comparisons between long-context backbones and external-memory architectures.

\begin{figure*}[!t]
    \centering
    \includegraphics[width=0.99\textwidth]{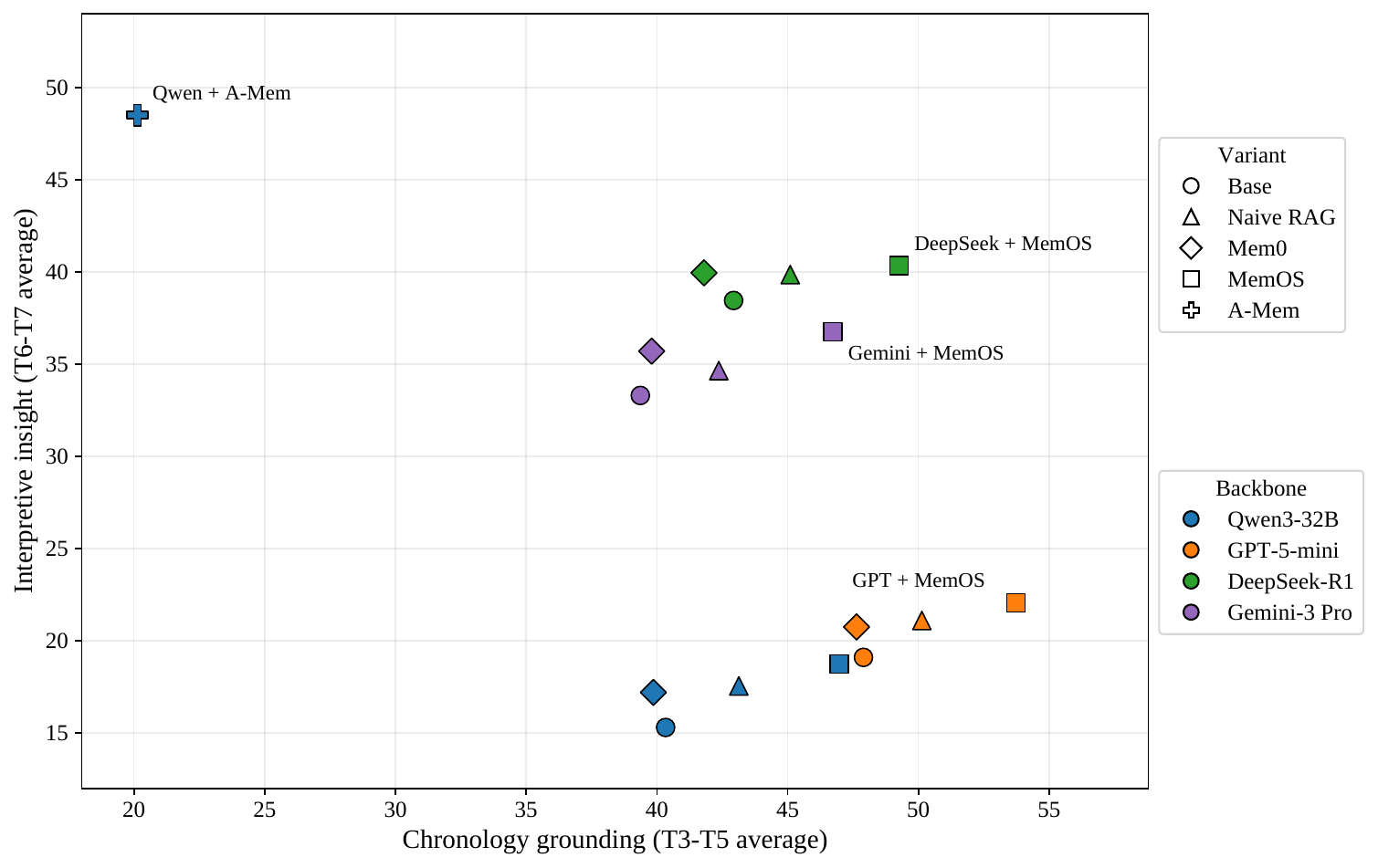}
    \caption{Grounding--insight map across evaluated systems. The x-axis averages the chronology-sensitive tasks $T_3$--$T_5$, and the y-axis averages the interpretive tasks $T_6$--$T_7$. Colors denote backbone families and marker shapes denote system variants. To keep the plot legible, Level-I recall deltas are reported in Table~\ref{tab:main_results} rather than encoded by marker size. The rightward frontier is populated by MemOS variants, while Qwen3-32B + A-Mem remains a high-insight but weak-grounding outlier.}
    \label{fig:tradeoff_map}
\end{figure*}

\begin{table*}[!t]
    \centering
    \scriptsize
    \setlength{\tabcolsep}{2.6pt}
    \renewcommand{\arraystretch}{1.07}
    \begin{tabular}{lccccccc}
        \toprule
        \rowcolor{gray!12}
        \multirow{2}{*}{\textbf{System}} & \multicolumn{3}{c}{\textbf{Level I: Fact \& Entity}} & \multicolumn{2}{c}{\textbf{Level II: Narrative Logic}} & \multicolumn{2}{c}{\textbf{Level III: Insight}} \\
        \cmidrule(lr){2-4} \cmidrule(lr){5-6} \cmidrule(lr){7-8}
        & \shortstack{\textbf{T1}\\Extract} & \shortstack{\textbf{T2}\\Abstain} & \shortstack{\textbf{T3}\\Temporal} & \shortstack{\textbf{T4}\\Order} & \shortstack{\textbf{T5}\\Trigger} & \shortstack{\textbf{T6}\\Mind--Body} & \shortstack{\textbf{T7}\\Principle} \\
        \midrule
        \rowcolor{gray!8} \multicolumn{8}{l}{\textbf{Backbone: Qwen3-32B}} \\
        Base & 59.9 & 66.0 & 44.4 & 40.5 & 36.1 & 14.3 & 16.3 \\
        + Naive RAG & 68.7{\scriptsize\,\textcolor{gray}{(+8.8)}} & 70.8{\scriptsize\,\textcolor{gray}{(+4.8)}} & 48.3{\scriptsize\,\textcolor{gray}{(+3.9)}} & 42.3{\scriptsize\,\textcolor{gray}{(+1.8)}} & 38.8{\scriptsize\,\textcolor{gray}{(+2.7)}} & 17.1{\scriptsize\,\textcolor{gray}{(+2.8)}} & 18.0{\scriptsize\,\textcolor{gray}{(+1.7)}} \\
        + Mem0 & \textbf{70.4}{\scriptsize\,\textcolor{gray}{(+10.5)}} & \textbf{75.2}{\scriptsize\,\textcolor{gray}{(+9.2)}} & 41.3{\scriptsize\,\textcolor{gray}{(-3.1)}} & 42.4{\scriptsize\,\textcolor{gray}{(+1.9)}} & 35.9{\scriptsize\,\textcolor{gray}{(-0.2)}} & 16.9{\scriptsize\,\textcolor{gray}{(+2.6)}} & 17.5{\scriptsize\,\textcolor{gray}{(+1.2)}} \\
        + MemOS & 64.4{\scriptsize\,\textcolor{gray}{(+4.5)}} & 72.4{\scriptsize\,\textcolor{gray}{(+6.4)}} & \textbf{52.7}{\scriptsize\,\textcolor{gray}{(+8.3)}} & \textbf{47.1}{\scriptsize\,\textcolor{gray}{(+6.6)}} & \textbf{41.1}{\scriptsize\,\textcolor{gray}{(+5.0)}} & 18.2{\scriptsize\,\textcolor{gray}{(+3.9)}} & 19.3{\scriptsize\,\textcolor{gray}{(+3.0)}} \\
        + A-Mem & 55.8{\scriptsize\,\textcolor{gray}{(-4.1)}} & 73.4{\scriptsize\,\textcolor{gray}{(+7.4)}} & 16.0{\scriptsize\,\textcolor{gray}{(-28.4)}} & 16.5{\scriptsize\,\textcolor{gray}{(-24.0)}} & 27.9{\scriptsize\,\textcolor{gray}{(-8.2)}} & \textbf{55.7}{\scriptsize\,\textcolor{gray}{(+41.4)}} & \textbf{41.3}{\scriptsize\,\textcolor{gray}{(+25.0)}} \\
        \addlinespace[2pt]
        \rowcolor{gray!8} \multicolumn{8}{l}{\textbf{Backbone: GPT-5-mini}} \\
        Base & 65.4 & 71.5 & 54.1 & 47.3 & 42.3 & 18.6 & 19.6 \\
        + Naive RAG & 72.2{\scriptsize\,\textcolor{gray}{(+6.8)}} & 75.5{\scriptsize\,\textcolor{gray}{(+4.0)}} & 57.3{\scriptsize\,\textcolor{gray}{(+3.2)}} & 48.8{\scriptsize\,\textcolor{gray}{(+1.5)}} & 44.3{\scriptsize\,\textcolor{gray}{(+2.0)}} & 21.3{\scriptsize\,\textcolor{gray}{(+2.7)}} & 20.9{\scriptsize\,\textcolor{gray}{(+1.3)}} \\
        + Mem0 & \textbf{73.2}{\scriptsize\,\textcolor{gray}{(+7.8)}} & \textbf{78.7}{\scriptsize\,\textcolor{gray}{(+7.2)}} & 51.6{\scriptsize\,\textcolor{gray}{(-2.5)}} & 48.7{\scriptsize\,\textcolor{gray}{(+1.4)}} & 42.6{\scriptsize\,\textcolor{gray}{(+0.3)}} & 20.7{\scriptsize\,\textcolor{gray}{(+2.1)}} & 20.8{\scriptsize\,\textcolor{gray}{(+1.2)}} \\
        + MemOS & 69.6{\scriptsize\,\textcolor{gray}{(+4.2)}} & 76.6{\scriptsize\,\textcolor{gray}{(+5.1)}} & \textbf{60.8}{\scriptsize\,\textcolor{gray}{(+6.7)}} & \textbf{52.9}{\scriptsize\,\textcolor{gray}{(+5.6)}} & \textbf{47.5}{\scriptsize\,\textcolor{gray}{(+5.2)}} & \textbf{21.5}{\scriptsize\,\textcolor{gray}{(+2.9)}} & \textbf{22.6}{\scriptsize\,\textcolor{gray}{(+3.0)}} \\
        \addlinespace[2pt]
        \rowcolor{gray!8} \multicolumn{8}{l}{\textbf{Backbone: DeepSeek-R1}} \\
        Base & 51.2 & 42.2 & 52.7 & 37.0 & 39.1 & 41.6 & 35.3 \\
        + Naive RAG & 57.7{\scriptsize\,\textcolor{gray}{(+6.5)}} & 47.9{\scriptsize\,\textcolor{gray}{(+5.7)}} & 57.0{\scriptsize\,\textcolor{gray}{(+4.3)}} & 38.0{\scriptsize\,\textcolor{gray}{(+1.0)}} & 40.3{\scriptsize\,\textcolor{gray}{(+1.2)}} & \textbf{43.4}{\scriptsize\,\textcolor{gray}{(+1.8)}} & 36.3{\scriptsize\,\textcolor{gray}{(+1.0)}} \\
        + Mem0 & \textbf{59.1}{\scriptsize\,\textcolor{gray}{(+7.9)}} & 51.1{\scriptsize\,\textcolor{gray}{(+8.9)}} & 50.5{\scriptsize\,\textcolor{gray}{(-2.2)}} & 37.0{\scriptsize\,\textcolor{gray}{(+0.0)}} & 37.9{\scriptsize\,\textcolor{gray}{(-1.2)}} & 43.1{\scriptsize\,\textcolor{gray}{(+1.5)}} & 36.8{\scriptsize\,\textcolor{gray}{(+1.5)}} \\
        + MemOS & 53.4{\scriptsize\,\textcolor{gray}{(+2.2)}} & \textbf{51.5}{\scriptsize\,\textcolor{gray}{(+9.3)}} & \textbf{61.1}{\scriptsize\,\textcolor{gray}{(+8.4)}} & \textbf{43.4}{\scriptsize\,\textcolor{gray}{(+6.4)}} & \textbf{43.3}{\scriptsize\,\textcolor{gray}{(+4.2)}} & 43.2{\scriptsize\,\textcolor{gray}{(+1.6)}} & \textbf{37.5}{\scriptsize\,\textcolor{gray}{(+2.2)}} \\
        \addlinespace[2pt]
        \rowcolor{gray!8} \multicolumn{8}{l}{\textbf{Backbone: Gemini-3 Pro}} \\
        Base & 50.5 & 52.2 & 45.3 & 36.0 & 36.8 & 35.4 & 31.2 \\
        + Naive RAG & 59.0{\scriptsize\,\textcolor{gray}{(+8.5)}} & 57.7{\scriptsize\,\textcolor{gray}{(+5.5)}} & 49.6{\scriptsize\,\textcolor{gray}{(+4.3)}} & 38.7{\scriptsize\,\textcolor{gray}{(+2.7)}} & 38.8{\scriptsize\,\textcolor{gray}{(+2.0)}} & 37.0{\scriptsize\,\textcolor{gray}{(+1.6)}} & 32.3{\scriptsize\,\textcolor{gray}{(+1.1)}} \\
        + Mem0 & \textbf{60.8}{\scriptsize\,\textcolor{gray}{(+10.3)}} & \textbf{62.3}{\scriptsize\,\textcolor{gray}{(+10.1)}} & 43.7{\scriptsize\,\textcolor{gray}{(-1.6)}} & 37.2{\scriptsize\,\textcolor{gray}{(+1.2)}} & 38.5{\scriptsize\,\textcolor{gray}{(+1.7)}} & \textbf{39.0}{\scriptsize\,\textcolor{gray}{(+3.6)}} & 32.4{\scriptsize\,\textcolor{gray}{(+1.2)}} \\
        + MemOS & 56.3{\scriptsize\,\textcolor{gray}{(+5.8)}} & 60.5{\scriptsize\,\textcolor{gray}{(+8.3)}} & \textbf{53.4}{\scriptsize\,\textcolor{gray}{(+8.1)}} & \textbf{44.1}{\scriptsize\,\textcolor{gray}{(+8.1)}} & \textbf{42.7}{\scriptsize\,\textcolor{gray}{(+5.9)}} & 37.9{\scriptsize\,\textcolor{gray}{(+2.5)}} & \textbf{35.6}{\scriptsize\,\textcolor{gray}{(+4.4)}} \\
        \bottomrule
    \end{tabular}
    \caption{Overall results on KnowMe-Bench across all evaluated systems. Parenthetical deltas are measured against the Base model within the same backbone family, and boldface marks the strongest absolute score within each backbone block. DeepSeek-R1 and Gemini-3 Pro follow the same Base/Naive-RAG/Mem0/MemOS protocol as the two primary backbones, while Qwen3-32B additionally includes A-Mem as a reflective-memory comparison. The pattern is consistent across families: retrieval-oriented variants help the fact-heavy tasks, whereas MemOS more reliably improves chronology-sensitive and insight-oriented tasks.}
    \label{tab:main_results}
\end{table*}

\subsection{Main Results}

Table~\ref{tab:main_results} reports the complete overall task-level results for all evaluated systems, while Appendix~\ref{app:extra_results} retains the per-dataset breakdowns and the backbone-specific radar views. Figure~\ref{fig:tradeoff_map} expands the core main-paper visualization to a full-width grounding--insight map; Level-I recall deltas are kept in Table~\ref{tab:main_results} so that the figure can remain readable.

\subsection{In-Depth Analysis \& Discussion}

\paragraph{Takeaway 1: Chronological logging is the decisive ingredient in non-linear narratives.}
The dataset-level views in Appendix~\ref{app:extra_results} expose the clearest failure mode on the flashback-heavy corpus: state-updating memory improves entity-centric retrieval but can mis-handle recalled material as a present-state overwrite. On Qwen3-32B in Dataset~1, Mem0 raises $T_1$/$T_2$ yet lowers $T_3$ by 3.5 points, whereas MemOS improves $T_3$/$T_4$ by 10.4/10.8. This pattern is consistent with an ``update paradox'' in which statements such as ``I liked apples as a child'' are absorbed as current state unless the memory store preserves chronological provenance.

\paragraph{Takeaway 2: Retrieval gains and interpretive gains are separable.}
Across backbones, Naive RAG and Mem0 remain strongest on the fact-heavy tasks, but they do not deliver the same gains on chronology-sensitive and interpretive tasks. Table~\ref{tab:main_results} shows this split repeatedly: on GPT-5-mini, Mem0 is best on $T_1$/$T_2$ at 73.2/78.7, whereas MemOS is best on $T_3$--$T_7$; Gemini-3 Pro exhibits the same pattern. A-Mem makes the contrast even sharper on Qwen3-32B, reaching 55.7/41.3 on $T_6$/$T_7$ while collapsing to 16.0/16.5 on $T_3$/$T_4$. The benchmark therefore separates at least two partially orthogonal capabilities: precise recall of explicit facts and stable reasoning over temporally grounded personal experience.

\paragraph{Takeaway 3: Stronger reasoning backbones still need explicit temporal scaffolding.}
The expanded evaluation on DeepSeek-R1 and Gemini-3 Pro shows that the central effect is architectural rather than backbone-specific. Both models improve under MemOS on $T_3$--$T_5$, even when retrieval-oriented alternatives remain competitive on $T_1$/$T_2$. Closed-book contamination checks reinforce this interpretation: on Level~II logic items, accuracy drops from 89.1 with original entity names to 33.2 after de-identification, while Level~III insight remains at 0 in both settings. The hardest items are therefore not explained by memorized literary knowledge alone; they continue to require evidence-grounded reasoning over the reconstructed chronological stream.

\section{Conclusion}
In this work, we introduced \textbf{KnowMe-Bench}, a benchmark designed to shift the evaluation of lifelong digital companions from simple fact retrieval to evidence-grounded person understanding. By leveraging high-density autobiographical narratives rather than sparse chat logs, we build a substrate that preserves the ``micro-texture'' of human experience---actions, inner thoughts, and environmental context---while remaining auditable at the level of aligned narrative evidence.

Our experiments reveal a clear evaluation gap in current long-horizon memory research. Retrieval-augmented baselines and entity-tracking systems improve factual recall, but they remain structurally fragile when the benchmark requires flashback-aware chronology, trigger--event linkage, and longer-range interpretive reasoning. The resulting update paradox shows why a personal memory system cannot be treated as a static fact database: without explicit temporal provenance, recalled past experience is too easily conflated with present state.

KnowMe-Bench provides a concrete testbed for this distinction through flashback-aware reconstruction, evidence-linked evaluation, and diagnostic comparisons across memory architectures. We hope it encourages future work to move beyond context-window extension and vector similarity toward memory systems that support stronger temporal grounding, more reliable longitudinal user modeling, and evidence-grounded reasoning over lived experience.

\section*{Limitations}
Methodologically, the benchmark must navigate the inherent subjectivity of literary analysis through a rigorous "LLM-as-a-Judge" protocol validated by human experts, while bearing the cost and operational complexity of a multi-agent generation and de-identification pipeline for dense autobiographical data.

\section*{Ethical considerations}

We strictly adhere to the licenses and usage policies of the open-source models and datasets utilized in our experiments. Our benchmark does not introduce additional risks regarding data privacy or human rights violations.

\bibliography{custom}

\newpage

\appendix


\clearpage
\onecolumn

\section{Faithfulness Verification Protocol}
\label{app:verification}

To operationalize the ``Faithfulness-First'' principle, we implement a generic verification layer that guards all generative transformations (Modules B, C, and D). This appendix details the computation of the semantic divergence score ($\delta$), the threshold configurations, and the specific prompts used for the Consistency Check Agent.

\subsection{Semantic Divergence Metric (\texorpdfstring{$\delta$}{delta})}

We define semantic divergence $\delta$ not merely as vector distance, but as a measure of \emph{propositional mismatch}. We employ a \textbf{Key Information Extraction (KIE)} overlap method.

Let $S$ be the source narrative segment and $T$ be the generated output (e.g., extracted ANUs or instantiated text). We prompt a Validator Agent to extract the set of atomic facts $\mathcal{F}(\cdot)$ from both texts (including entities, timestamps, and actions).

The divergence score is computed as a weighted combination of \emph{Omission Rate} ($\delta_{miss}$) and \emph{Hallucination Rate} ($\delta_{hall}$):

\begin{equation}
    \delta(S, T) = \alpha \cdot \underbrace{\left( 1 - \frac{|\mathcal{F}(S) \cap \mathcal{F}(T)|}{|\mathcal{F}(S)|} \right)}_{\delta_{miss}} + \beta \cdot \underbrace{\left( \frac{|\mathcal{F}(T) \setminus \mathcal{F}(S)|}{|\mathcal{F}(T)|} \right)}_{\delta_{hall}}
\end{equation}

where:
\begin{itemize}
    \item $\mathcal{F}(S) \cap \mathcal{F}(T)$ represents facts present in both source and output.
    \item $\mathcal{F}(T) \setminus \mathcal{F}(S)$ represents new facts introduced in the output (hallucinations).
    \item We set $\alpha=0.4$ and $\beta=0.6$, penalizing hallucinations more strictly than minor omissions to prevent corruption of the ground truth.
\end{itemize}

\subsection{Threshold Configuration (\texorpdfstring{$\epsilon$}{epsilon})}

The acceptance threshold $\epsilon$ varies by module sensitivity:
\begin{table}[!h]
\centering
\small
\setlength{\tabcolsep}{6pt}        
\renewcommand{\arraystretch}{1.2} 

\begin{tabular}{l|c|l}
\hline
\textbf{Module} & \textbf{Threshold (\texorpdfstring{$\epsilon$}{epsilon})} & \textbf{Rationale} \\
\hline
Mod B (Extraction) & $0.05$ & High tolerance for stylistic compression, zero tolerance for entity loss. \\
Mod C (Realignment) & $0.00$ & Strict logic check; timestamp order must match the causal graph exactly. \\
Mod D (Instantiation) & $0.03$ & Allows minor grammatical changes for first-person flow, but bans new adjectives. \\
\hline
\end{tabular}

\caption{Divergence thresholds for automatic revision triggers.}
\label{tab:thresholds}
\end{table}

\subsection{Prompt Implementation}

Below is the specific system prompt used by the \textbf{Consistency Check Agent} to evaluate Module B (ANU Extraction). This prompt enforces the calculation of $\delta$ through step-by-step verification.

\begin{center}
\fbox{\begin{minipage}{0.95\textwidth}
\smallra
\textbf{System Prompt: The Auditor}

You are a strict Data Auditor. Your task is to compare the \texttt{SOURCE\_TEXT} against the extracted \texttt{ANU\_JSON}.

\textbf{Step 1: Fact Extraction}
List all atomic facts in \texttt{SOURCE\_TEXT} (Entities, Actions, Time, Location).
List all atomic facts represented in \texttt{ANU\_JSON}.

\textbf{Step 2: Discrepancy Analysis}
Identify two types of errors:
1. \textbf{[MISSING]}: A critical fact (e.g., a name ``John'', a time ``noon'') exists in Source but is absent in JSON.
2. \textbf{[HALLUCINATION]}: A fact exists in JSON but is NOT supported by Source (e.g., adding an adjective ``angry'' when the text only said ``said'').

\textbf{Step 3: Verification Decision}
If there are ANY [HALLUCINATION] tags or significant [MISSING] tags, return Status: \textsc{REJECT}.
Otherwise, return Status: \textsc{PASS}.

\textbf{Output Format:}
\{
  "status": "PASS" | "REJECT",
  "score": [0.0 - 1.0],
  "feedback": "Specific instructions on what to fix..."
\}
\end{minipage}}
\end{center}

\subsection{Error Feedback Loop}\label{app:verification-feedback}

If $\delta(S, T) > \epsilon$, the system enters a \emph{Revision Loop}:
\begin{enumerate}
    \item The Validator Agent generates a natural language feedback message $M_{fb}$ (e.g., \emph{``Error: You missed the location `gas station' mentioned in line 3.''}).
    \item The Generator Agent receives the history $[S, T_{old}, M_{fb}]$ and attempts a regeneration $T_{new}$.
    \item This loop repeats up to $k_{max}=3$ times. If convergence fails, the sample is flagged for manual human review.
\end{enumerate}

\noindent
For the mnestic realignment module, we use the following action semantics:
\begin{itemize}
    \item \textbf{MAINTAIN:} Extends the current timeline.
    \item \textbf{PUSH($t_{new}$):} Triggered by \textbf{Structural Narrative Inversions} (assigning $C_{event}$ to $t_{new}$). It pushes a new layer for sustained flashbacks.
    \item \textbf{POP():} Returns to the parent layer's active timestamp after the recollection ends.
    \item \textbf{TRANSIENT:} Marks fleeting \textbf{Associative Triggers} ($T_{trigger}$) that evoke a memory without altering the stack structure.
\end{itemize}

\clearpage
\section{Examples}
\label{sec:appendix}

\begin{prettybox}{Example I: ANU Extraction}

    \begin{tcolorbox}[blanker, left=10pt, borderline west={3pt}{0pt}{gray!30}]
        \textbf{\textit{Input Segment:}} \\
        ``I put the coffee cup on the windowsill. Rain is still hitting the glass.''
    \end{tcolorbox}

    \vspace{0.2cm}

    \noindent
    \begin{tabular}{@{}ll@{}}
        \tagbox{ID} & \textbf{ANU-001} \\[3pt]
        \tagbox{Time Anchor} & Morning, before rain stops \\[3pt]
        \tagbox{Location} & Windowsill
    \end{tabular}

    \vspace{0.2cm}

    \begin{tcolorbox}[colback=accentbg, frame hidden, arc=3pt, boxsep=1pt, left=6pt, right=6pt, top=5pt, bottom=5pt]
        \noindent \textbf{Content Details:}

        \vspace{2pt}
        {\color{linecolor}\hrule height 0.8pt}
        \vspace{4pt}

        \inlineitem{Action}{I place the coffee cup on the windowsill.}
        \inlineitem{Environment}{Rain is still hitting the glass.}

        \vspace{1pt}
        \noindent
        {\color{gray}\scriptsize
        \textbf{Dialogue:} \textit{None} \quad $\cdot$ \quad \textbf{Mind:} \textit{None}}
    \end{tcolorbox}

\end{prettybox}

\begin{prettybox}{Example II: Final Data Instance}

    \noindent
    \begin{tabular}{@{}ll@{}}
        \tagbox{ID} & \textbf{101} \\[3pt]
        \tagbox{Timestamp} & 1966-04-25 19:00:00 \\[3pt]
        \tagbox{Location} & Windowsill
    \end{tabular}

    \vspace{0.2cm}

    \begin{tcolorbox}[colback=accentbg, frame hidden, arc=3pt, boxsep=1pt, left=6pt, right=6pt, top=5pt, bottom=5pt]
        \noindent \textbf{Content Details:}

        \vspace{2pt}
        {\color{linecolor}\hrule height 0.8pt}
        \vspace{4pt}

        \inlineitem{Action}{I place the coffee cup on the windowsill.}
        \inlineitem{Environment}{I see rain is still hitting the glass.}

        \vspace{1pt}
        \noindent
        {\color{gray}\scriptsize
        \textbf{Dialogue:} \textit{None} \quad $\cdot$ \quad \textbf{Mind:} \textit{None} \quad $\cdot$ \quad \textbf{Background:} \textit{None}}
    \end{tcolorbox}

\end{prettybox}

Unless otherwise stated, the default strict acceptance threshold is $\epsilon = 0.03$; Module B uses $0.05$ as listed in Table~\ref{tab:thresholds}.

We applied a rigorous Context-Aware De-identification Pipeline. Key entities were mapped to consistent pseudonyms (e.g., ``Elena'' $\to$ ``Subject\_A'') to preserve coreference chains, and geolocation markers were coarsened to ensure no residual PII remained.



\clearpage
\onecolumn

\newcommand{\abstain}{\texttt{ABSTAIN}}

\newtcolorbox{promptcard}[2][]{%
  enhanced,
  colback=white,
  colframe=maincolor,
  fonttitle=\bfseries\large,
  title={#2},
  boxed title style={colback=maincolor, frame hidden, arc=3pt},
  attach boxed title to top left={yshift=-10pt, xshift=10pt},
  boxrule=1.0pt,
  arc=3pt,
  top=10pt,bottom=8pt,left=10pt,right=10pt,
  before upper=\normalsize,
  #1
}

\newtcolorbox{promptcardhalf}[2][]{%
  enhanced,
  colback=white,
  colframe=maincolor,
  fonttitle=\bfseries\large,
  title={#2},
  boxed title style={colback=maincolor, frame hidden, arc=3pt},
  attach boxed title to top left={yshift=-10pt, xshift=10pt},
  boxrule=1.0pt,
  arc=3pt,
  top=12pt,bottom=10pt,left=10pt,right=10pt,
  before upper=\normalsize,
  valign=top,
  height=0.40\textheight, 
  #1
}

\newcommand{\pcsubtitle}[1]{}

\newcommand{\pcsep}{\vspace{0.35em}}
\newcommand{\pcfield}[2]{\textbf{#1} & #2\\}

\newenvironment{compactitem}{%
  \begin{itemize}
    \setlength{\itemsep}{1pt}
    \setlength{\topsep}{1pt}
    \setlength{\parskip}{0pt}
}{\end{itemize}}

\lstdefinestyle{pcskeleton}{%
  basicstyle=\ttfamily\footnotesize,
  breaklines=true,
  breakatwhitespace=false,
  columns=fullflexible,
  keepspaces=true,
  showstringspaces=false
}
\newtcblisting{skeletonbox}{%
  listing only,
  colback=accentbg,
  colframe=maincolor!35,
  boxrule=0.5pt,
  arc=2pt,
  left=6pt,right=6pt,top=4pt,bottom=4pt,
  listing options={style=pcskeleton}
}

\begingroup
\raggedbottom
\section{Prompt Cards}
\label{app:prompt-cards}
\enlargethispage{3\baselineskip}
\vspace{-0.6em}

We summarize the prompt files via \emph{Prompt Cards} to improve auditability while avoiding full prompt dumps.
Each card reports a minimal contract: \textbf{Role}, \textbf{Inputs}, \textbf{Output Contract}, \textbf{Reject/Gate},
and \textbf{Hard Constraints}. Redundant boilerplate and in-context examples are omitted.
\textit{Note: The cards intentionally abstract the original prompts by removing boilerplate and in-context examples; full prompt texts are provided in the supplementary material.}
\vspace{-0.7em}

\subsection{Data Construction Pipeline (Modules A--D)}
\label{app:prompt-cards-pipeline}
\vspace{-0.5em}

\begin{promptcard}{A. Segmentation}

\begin{tabular}{@{}p{0.15\linewidth}p{0.83\linewidth}@{}}
\pcfield{Role}{Slice raw narrative into segments without altering any character.}
\pcfield{Inputs}{Raw narrative text \texttt{N}.}
\pcfield{Output}{List of segment records with \texttt{segment\_id}, \texttt{start\_index}, \texttt{end\_index}, \texttt{text} (verbatim substring).}
\pcfield{Gate}{None (extractive slicing only).}
\end{tabular}

\pcsep
\textbf{Hard Constraints.}
\begin{compactitem}
  \item \textbf{Verbatim preservation}: no rewriting/summarization/deletion; slicing is index-based only.
  \item \textbf{Semantic boundary}: cut at scene/event/time/location shifts; do not break sentences or ongoing dialogue.
  \item \textbf{Length (prompt-level guidance)}: keep segments approximately within the token budget specified in the prompt.
\end{compactitem}

\pcsep
\textbf{Skeleton.}
\begin{skeletonbox}
Input: raw narrative N.
Operation: boundary-based index slicing only (verbatim).
Output(JSON): [{segment_id, start_index, end_index, text}, ...]
\end{skeletonbox}
\end{promptcard}

\vspace{0.25em}

\begin{promptcard}{B. ANU + Check}

\begin{tabular}{@{}p{0.15\linewidth}p{0.83\linewidth}@{}}
\pcfield{Role}{Extract Atomic Narrative Units (ANUs) and audit against the source span for omission/hallucination.}
\pcfield{Inputs}{One segment from Module A (verbatim).}
\pcfield{Output}{(B) ANU list with \texttt{id}, \texttt{t\_anchor} (verbatim), \texttt{location} (required),
\texttt{content\{action,dialogue,environment,background,mind\}}.
(B-check) verdict \texttt{\{semantic\_difference\_score,status,issues\}}.}
\pcfield{Gate}{Reject if $\delta > 0.05$ (information loss or hallucination).}
\end{tabular}

\pcsep
\textbf{Hard Constraints.}
\begin{compactitem}
  \item \textbf{Granularity}: $\leq$ 3 physical actions \emph{or} $\leq$ 3 dialogue turns per ANU; otherwise split.
  \item \textbf{No abstract state}: prohibit vague mental labels; decompose into explicit micro-behaviors and/or explicit \texttt{mind}.
  \item \textbf{Spatiotemporal unity}: space change or noticeable time jump triggers a new ANU.
\end{compactitem}

\pcsep
\textbf{Skeleton.}
\begin{skeletonbox}
Input: one segment (verbatim).
Output: ANU list with mandatory location + five primitives.
Gate: run audit; REJECT if delta > 0.05 or hallucination detected.
\end{skeletonbox}
\end{promptcard}

\clearpage
\enlargethispage{2\baselineskip}

\begin{promptcard}{C. Timeline + Check}

\begin{tabular}{@{}p{0.15\linewidth}p{0.83\linewidth}@{}}
\pcfield{Role}{Maintain a stack-based mnestic realignment state machine and assign chronological placements.}
\pcfield{Inputs}{Current ANU; lookahead (next 3 ANUs); current stack time; stack depth.}
\pcfield{Output}{(C) \texttt{\{action,time\_value,reasoning\}} with \texttt{action}\,$\in$\,\{MAINTAIN,PUSH,POP,TRANSIENT\}.
(C-check) \texttt{\{status,logic\_error,correction\_suggestion\}}.}
\pcfield{Gate}{C-check rejects boundary misalignment or implausible duration allocation.}
\end{tabular}

\pcsep
\textbf{Hard Constraints.}
\begin{compactitem}
  \item \textbf{Lookahead-based scope}: distinguish transient triggers vs sustained flashbacks.
  \item \textbf{State discipline}: PUSH only if subsequent ANUs belong to past; POP on return; TRANSIENT if immediate return next ANU.
  \item \textbf{C-check}: (i) duration plausibility; (ii) PUSH must be justified by immediately subsequent content.
\end{compactitem}

\pcsep
\textbf{Skeleton.}
\begin{skeletonbox}
Inputs: current ANU; lookahead(next 3); stack time; stack depth.
Decide: MAINTAIN / PUSH / POP / TRANSIENT.
Output: {action, time_value(YYYY-MM-DD HH:MM:SS), reasoning}.
\end{skeletonbox}
\end{promptcard}

\vspace{0.25em}

\begin{promptcard}{D. Narrative + Check}

\begin{tabular}{@{}p{0.15\linewidth}p{0.83\linewidth}@{}}
\pcfield{Role}{Instantiate each aligned ANU into first-person experience; reject forbidden distortions.}
\pcfield{Inputs}{Chronologically aligned ANU with optional fields \texttt{action/dialogue/environment/background/mind}.}
\pcfield{Output}{(D) one first-person paragraph. (D-check) \texttt{\{status,hallucination\_detected,details\}}.}
\pcfield{Gate}{Reject if $\delta>0.03$ or if any embellishment/emotional injection is detected.}
\end{tabular}

\pcsep
\textbf{Hard Constraints.}
\begin{compactitem}
  \item \textbf{Component-wise subjectivization}: translate each present field into immediate ``I''-perspective experience.
  \item \textbf{Strict coverage}: cover all fields present (no omission).
  \item \textbf{No hallucination}: do not add adjectives/emotions absent from \texttt{mind}/\texttt{environment}.
\end{compactitem}

\pcsep
\textbf{Skeleton.}
\begin{skeletonbox}
Method: component-wise subjectivization (I-perspective).
Gate: REJECT if delta > 0.03 or distortion detected.
\end{skeletonbox}
\end{promptcard}

\clearpage
\enlargethispage{2\baselineskip}

\subsection{Benchmark Instance Generation (Level I: T1--T3)}
\label{app:prompt-cards-level1}

\begin{promptcardhalf}{E. T1: Extraction}

\begin{tabular}{@{}p{0.15\linewidth}p{0.83\linewidth}@{}}
\pcfield{Role}{Generate retrieval-focused QA with explicit spatiotemporal constraints.}
\pcfield{Inputs}{Evidence items with \texttt{id,timestamp,location,category,related\_time?}.}
\pcfield{Output}{\texttt{[{id, question, answer, evidence\_ids}, ...]}.}
\pcfield{Gate}{N/A.}
\end{tabular}

\pcsep
\textbf{Hard Constraints.}
\begin{compactitem}
  \item \textbf{Uniqueness}: include sufficient constraints so the answer is unique.
  \item \textbf{Evidence anchoring}: \texttt{evidence\_ids} must point to the minimal supporting IDs.
\end{compactitem}

\pcsep
\textbf{Skeleton.}
\begin{skeletonbox}
Inputs: evidence items with time + location anchors.
Produce: one uniquely answerable question + minimal evidence_ids.
Output(JSON): [{id, question, answer, evidence_ids}, ...]
\end{skeletonbox}
\end{promptcardhalf}

\vspace{0.8em} 

\begin{promptcardhalf}{F. T2: Abstention}

\begin{tabular}{@{}p{0.15\linewidth}p{0.83\linewidth}@{}}
\pcfield{Role}{Compose true fragments into a false relation to test abstention (anti-hallucination).}
\pcfield{Inputs}{Same evidence schema as T1.}
\pcfield{Output}{\texttt{[{id, question, answer, evidence\_ids}, ...]}.}
\pcfield{Gate}{Answer must be the fixed abstention token \texttt{ABSTAIN}.}
\end{tabular}

\pcsep
\textbf{Hard Constraints.}
\begin{compactitem}
  \item \textbf{Entity validity, relation invalidity}: all entities must exist in evidence, but their relations must be wrong.
  \item \textbf{Trap strategies}: entity swapping; spatiotemporal distortion; false causality via unrelated true anchors.
\end{compactitem}

\pcsep
\textbf{Skeleton.}
\begin{skeletonbox}
Inputs: valid entities/time/location anchors from evidence.
Construct: mismatching relation while keeping anchors individually true.
Gate: answer MUST be ABSTAIN (fixed token).
\end{skeletonbox}
\end{promptcardhalf}

\clearpage

\begin{promptcard}{G. T3: Temporal}

\begin{tabular}{@{}p{0.15\linewidth}p{0.83\linewidth}@{}}
\pcfield{Role}{Generate duration computation and real-world ordering QA under non-linear narration.}
\pcfield{Inputs}{Evidence items with \texttt{timestamp} and optional \texttt{related\_time}.}
\pcfield{Output}{\texttt{[{id, question, answer, evidence\_ids}, ...]}.}
\pcfield{Gate}{N/A.}
\end{tabular}

\pcsep
\textbf{Hard Constraints.}
\begin{compactitem}
  \item \textbf{Duration}: answer = end timestamp $-$ start timestamp; include start/end anchors in \texttt{evidence\_ids}.
  \item \textbf{Ordering}: order by real-world occurrence time (not narrative order); do not leak explicit timestamps in options.
  \item \textbf{Trigger vs recalled content}: separate trigger at \texttt{timestamp} from recalled event at \texttt{related\_time}.
\end{compactitem}

\pcsep
\textbf{Skeleton.}
\begin{skeletonbox}
Duration: compute end\_ts - start\_ts; evidence\_ids include start+end.
Ordering: ask real-world order (no explicit timestamps in options).
Key: separate trigger(timestamp) vs recalled content(related\_time).
\end{skeletonbox}
\end{promptcard}


\enlargethispage{2\baselineskip}

\subsection{Benchmark Instance Generation (Level II: T4--T5)}
\label{app:prompt-cards-level2}

\begin{promptcard}[top=8pt,bottom=6pt]{H. T4: Logical Event Ordering}

\begin{tabular}{@{}p{0.15\linewidth}p{0.83\linewidth}@{}}
\pcfield{Role}{Generate QA that requires ordering discrete events by a \emph{non-temporal} semantic dimension (e.g., escalation of danger), rather than by explicit timestamps.}
\pcfield{Inputs}{A set of evidence items with \texttt{id,timestamp,location,category,related\_time?} and content fields (e.g., \texttt{action/dialogue/environment/background/mind}).}
\pcfield{Output}{\texttt{[{id, question, answer, evidence\_ids}, ...]}; \texttt{question} must specify the ordering \texttt{dimension} and require a ranked list (e.g., top-3); \texttt{answer} must provide the ordered events and brief justifications.}
\pcfield{Gate}{N/A.}
\end{tabular}

\pcsep
\textbf{Hard Constraints.}
\begin{compactitem}
  \item \textbf{Non-temporal ordering}: the rank criterion must not reduce to chronological order; explicitly state a semantic \texttt{dimension} (severity/risk/urgency/blame, etc.).
  \item \textbf{Event discreteness}: each ranked element must correspond to a concrete evidence-grounded event (not a diffuse theme).
  \item \textbf{Justified ranking}: provide short, evidence-based reasons for each ordering decision.
  \item \textbf{Evidence closure}: \texttt{evidence\_ids} must include all items necessary to recover the ranked events and the ordering rationale.
\end{compactitem}

\vspace{-0.15em}
\textbf{Skeleton.}
\begin{skeletonbox}
Inputs: evidence items.
Select: a coherent set of discrete events.
Choose: a non-temporal semantic ordering dimension.
Ask: rank events under the dimension (e.g., most->least severe).
Answer: ordered list + brief justifications; evidence_ids cover all ranked events.
\end{skeletonbox}
\end{promptcard}

\clearpage
\vspace*{\fill}

\begin{promptcard}{I. T5: Mnestic Trigger Analysis}

\begin{tabular}{@{}p{0.15\linewidth}p{0.83\linewidth}@{}}
\pcfield{Role}{Generate QA that identifies the sensory cue or associative trigger that evokes memory retrieval under stream-of-consciousness narration.}
\pcfield{Inputs}{Evidence items with \texttt{id,timestamp,location,category,related\_time?} and content fields (especially \texttt{environment/mind/background}).}
\pcfield{Output}{\texttt{[{id, question, answer, evidence\_ids}, ...]}; \texttt{question} must ask for the \emph{specific trigger} (object/sound/smell/visual cue/phrase) and link it to the recalled content; \texttt{answer} must name the trigger and explain the trigger$\rightarrow$memory linkage.}
\pcfield{Gate}{N/A.}
\end{tabular}

\pcsep
\textbf{Hard Constraints.}
\begin{compactitem}
  \item \textbf{Concrete trigger}: the asked trigger must be an explicitly stated, perceivable cue (not a generic mood).
  \item \textbf{Trigger-to-recall linkage}: the evidence must support a transition into recalled content (often via \texttt{related\_time} or an immediate shift in \texttt{mind}/\texttt{background}).
  \item \textbf{No invention}: do not introduce triggers or recalled details absent from evidence.
  \item \textbf{Minimal evidence}: \texttt{evidence\_ids} should include the trigger span and the recalled-content span.
\end{compactitem}

\pcsep
\textbf{Skeleton.}
\begin{skeletonbox}
Inputs: evidence items (timestamp + optional related_time).
Find: explicit sensory/associative trigger + ensuing recall/memory content.
Ask: which specific cue triggered the memory (and what it evoked).
Answer: name trigger + explain linkage; evidence_ids include trigger+recall spans.
Output(JSON): [{id, question, answer, evidence_ids}, ...]
\end{skeletonbox}
\end{promptcard}

\vspace*{\fill}


\clearpage
\begingroup
\raggedbottom

\subsection{Benchmark Instance Generation (Level III: T6--T7)}
\label{app:prompt-cards-level3}
\vspace{0.6em}

\noindent
\begin{promptcard}{J. T6: Mind-Body Interaction}

\begin{tabular}{@{}p{0.15\linewidth}p{0.83\linewidth}@{}}
\pcfield{Role}{Generate QA that explains the duality between external actions and internal states, including ironic or contradictory behaviors.}
\pcfield{Inputs}{Evidence items with \texttt{id,timestamp,location,category,related\_time?} and content fields, requiring both \texttt{action/dialogue} and \texttt{mind} (or equivalent internal cues).}
\pcfield{Output}{\texttt{[{id, question, answer, evidence\_ids}, ...]}; \texttt{question} must require relating bodily/behavioral manifestation to internal psychological state; \texttt{answer} must provide an evidence-grounded explanation of the interaction.}
\pcfield{Gate}{Reject if the selected evidence lacks either (i) observable external manifestation (\texttt{action/dialogue}) or (ii) internal-state cues (\texttt{mind} or implied cognition).}
\end{tabular}

\pcsep
\textbf{Hard Constraints.}
\begin{compactitem}
  \item \textbf{Dual-track requirement}: include both external behavior and internal state in the evidence and in the answer.
  \item \textbf{Interaction focus}: target explanation of irony/contradiction, mismatch, or expressive mapping (body/behavior as signal of mind).
  \item \textbf{Non-speculative}: do not add psychological claims beyond what can be justified by explicit narrative cues.
  \item \textbf{Evidence anchoring}: \texttt{evidence\_ids} must point to the minimal set supporting both tracks and their linkage.
\end{compactitem}

\pcsep
\textbf{Skeleton.}
\begin{skeletonbox}
Inputs: evidence items containing action/dialogue + mind cues.
Ask: how external manifestation reflects/contrasts internal state (incl. irony/contradiction).
Answer: cite external cues + internal cues + explain linkage.
Gate: REJECT if either track is missing.
Output(JSON): [{id, question, answer, evidence_ids}, ...]
\end{skeletonbox}
\end{promptcard}

\vspace{0.6em}

\noindent\textit{Why no Prompt Card for T7.}
Task~7 (\emph{Expert-Annotated Insight}) consists of open-ended questions curated by literary experts and is therefore not generated by an automatic prompt-based instance construction procedure. Accordingly, we do not provide a Prompt Card for T7. In evaluation, T7 is assessed under the Level~III rubric described in Section~\ref{sec:scoring-protocol}.

\vspace{0pt plus 1fill}

\endgroup
\endgroup


\clearpage
\begingroup
\raggedbottom

\def\QASectionToIntro{\vspace{0.55em}}
\def\QAIntroToLevel{\vspace{0.80em}}
\def\QALevelToCard{\vspace{0.8em}}
\def\QACardToCard{\vspace{0.8em}}

\section{QA Examples}
\label{app:qa-examples}
\QASectionToIntro

\noindent
We provide one representative QA instance for each task (T1--T7) from the released benchmark.
To ensure compatibility with memory systems that do not retain evidence IDs (or use remapped indices), we present \emph{QA-only} examples without evidence identifiers or anchor excerpts.
For T2, we normalize the reference abstention response to the fixed token \abstain\ for evaluation consistency.
\QAIntroToLevel

\subsection{Level I QA Examples (T1--T3)}
\label{app:qa-examples-level1}
\QALevelToCard

\begin{prettybox}{Example D.1: T1 Context-Aware Information Extraction}

    \begin{tcolorbox}[blanker, left=10pt, borderline west={3pt}{0pt}{gray!30}]
        \textbf{\textit{Question:}} \\
        ``On February 15, 1951, at the school on via Toledo, what specific items did Nunzia bring as gifts for Professor Oliviero?''
    \end{tcolorbox}

    \vspace{0.08cm}

    \noindent{\color{gray}\scriptsize
    \textbf{Task:} T1 \quad $\cdot$ \quad \textbf{Dataset:} Dataset 2}

    \vspace{0.10cm}

    \begin{tcolorbox}[colback=accentbg, frame hidden, arc=3pt, boxsep=1pt,
      left=6pt, right=6pt, top=4pt, bottom=4pt]
        \noindent \textbf{Answer:} Coffee and sugar.
    \end{tcolorbox}

\end{prettybox}

\QACardToCard

\begin{prettybox}{Example D.2: T2 Adversarial Abstention}

    \begin{tcolorbox}[blanker, left=10pt, borderline west={3pt}{0pt}{gray!30}]
        \textbf{\textit{Question:}} \\
        ``On July 27, 1963, while at the rentals, what did Stefano Carracci say to Pinocchia when he presented her with a small box containing a gold chain with a heart-shaped pendant?''
    \end{tcolorbox}

    \vspace{0.08cm}

    \noindent{\color{gray}\scriptsize
    \textbf{Task:} T2 \quad $\cdot$ \quad \textbf{Dataset:} Dataset 2}

    \vspace{0.10cm}

    \begin{tcolorbox}[colback=accentbg, frame hidden, arc=3pt, boxsep=1pt,
      left=6pt, right=6pt, top=4pt, bottom=4pt]
        \noindent \textbf{Answer:} \abstain

        \vspace{1pt}
        \noindent{\color{gray}\scriptsize
        \textbf{Note:} The released reference is an abstention-style response; we standardize it to the fixed token \abstain\ for scoring.}
    \end{tcolorbox}

\end{prettybox}

\QACardToCard

\begin{prettybox}{Example D.3: T3 Temporal Reasoning}

    \begin{tcolorbox}[blanker, left=10pt, borderline west={3pt}{0pt}{gray!30}]
        \textbf{\textit{Question:}} \\
        ``On October 20, 1950, how much time passed between the narrator being hit with a paper wad by Lina and Professor Oliviero stumbling and falling motionless to the floor?''
    \end{tcolorbox}

    \vspace{0.08cm}

    \noindent{\color{gray}\scriptsize
    \textbf{Task:} T3 \quad $\cdot$ \quad \textbf{Dataset:} Dataset 2}

    \vspace{0.10cm}

    \begin{tcolorbox}[colback=accentbg, frame hidden, arc=3pt, boxsep=1pt,
      left=6pt, right=6pt, top=4pt, bottom=4pt]
        \noindent \textbf{Answer:} 5 minutes.
    \end{tcolorbox}

\end{prettybox}

\clearpage

\subsection{Level II QA Examples (T4--T5)}
\label{app:qa-examples-level2}
\QALevelToCard

\begin{prettybox}{Example D.4: T4 Logical Event Ordering}

    \begin{tcolorbox}[blanker, left=10pt, borderline west={3pt}{0pt}{gray!30}]
        \textbf{\textit{Question:}} \\
        ``Arrange the following events based on the escalation of the girls' bravery and transgression in confronting their fears regarding Don Achille.''
    \end{tcolorbox}

    \vspace{0.08cm}

    \noindent{\color{gray}\scriptsize
    \textbf{Task:} T4 \quad $\cdot$ \quad \textbf{Dataset:} Dataset 2}

    \vspace{0.10cm}

    \begin{tcolorbox}[colback=accentbg, frame hidden, arc=3pt, boxsep=1pt,
      left=6pt, right=6pt, top=4pt, bottom=4pt]
        \noindent \textbf{Answer (increasing bravery/transgression):}
        \begin{compactitem}
          \item Throwing the dolls into the basement.
          \item Entering the building and climbing the dark staircase.
          \item Accusing Don Achille face-to-face at his door.
        \end{compactitem}
    \end{tcolorbox}

\end{prettybox}

\QACardToCard

\begin{prettybox}{Example D.5: T5 Mnestic Trigger Analysis}

    \begin{tcolorbox}[blanker, left=10pt, borderline west={3pt}{0pt}{gray!30}]
        \textbf{\textit{Question:}} \\
        ``The narrator is physically in the Staircase in May 1954, but her consciousness is mentally reliving a pivotal event from May 1953. What specific environmental anchor triggers this spatiotemporal jump and how does it restructure her current relationship with Lila?''
    \end{tcolorbox}

    \vspace{0.08cm}

    \noindent{\color{gray}\scriptsize
    \textbf{Task:} T5 \quad $\cdot$ \quad \textbf{Dataset:} Dataset 2}

    \vspace{0.10cm}

    \begin{tcolorbox}[colback=accentbg, frame hidden, arc=3pt, boxsep=1pt,
      left=6pt, right=6pt, top=4pt, bottom=4pt]
        \noindent \textbf{Answer:}
        The trigger is the dark staircase (together with the remembered purple light in the courtyard), which precipitates a chronological displacement back to the earlier episode and reframes the narrator's bond with Lila as grounded in that shared transgressive ascent.
    \end{tcolorbox}

\end{prettybox}


\clearpage

\subsection{Level III QA Examples (T6--T7)}
\label{app:qa-examples-level3}
\QALevelToCard

\begin{prettybox}{Example D.6: T6 Mind-Body Interaction}

    \begin{tcolorbox}[blanker, left=10pt, borderline west={3pt}{0pt}{gray!30}]
        \textbf{\textit{Question:}} \\
        ``The narrator physically wakes up and checks her leg, then decides to imitate Lina's walk. How does the simultaneous internal logic about her mother rationalize this behavior?''
    \end{tcolorbox}

    \vspace{0.08cm}

    \noindent{\color{gray}\scriptsize
    \textbf{Task:} T6 \quad $\cdot$ \quad \textbf{Dataset:} Dataset 2}

    \vspace{0.10cm}

    \begin{tcolorbox}[colback=accentbg, frame hidden, arc=3pt, boxsep=1pt,
      left=6pt, right=6pt, top=4pt, bottom=4pt]
        \noindent \textbf{Answer:}
        The narrator’s bodily checking and mimicry are justified by an internal fear of inheriting her mother’s lameness. By imitating Lina’s walk, she treats Lina’s bodily pattern as a protective counter-model, rationalizing the imitation as a way to ward off the threatened repetition of her mother’s gait.
    \end{tcolorbox}

\end{prettybox}

\QACardToCard

\begin{prettybox}{Example D.7: T7 Expert-Annotated Insight}

    \begin{tcolorbox}[blanker, left=10pt, borderline west={3pt}{0pt}{gray!30}]
        \textbf{\textit{Question:}} \\
        ``How do I view the conflict between `academic success' and `sense of belonging to the neighborhood'?''
    \end{tcolorbox}

    \vspace{0.08cm}

    \noindent{\color{gray}\scriptsize
    \textbf{Task:} T7 \quad $\cdot$ \quad \textbf{Dataset:} Dataset 2}

    \vspace{0.10cm}

    \begin{tcolorbox}[colback=accentbg, frame hidden, arc=3pt, boxsep=1pt,
      left=6pt, right=6pt, top=4pt, bottom=4pt]
        \noindent \textbf{Answer:}
        I view academic success as a primary route of escape and upward mobility, while simultaneously relying on the neighborhood’s interpersonal ties as a source of recognition that stabilizes my self-worth.

        \vspace{1pt}
        \noindent{\color{gray}\scriptsize
        \textbf{Note:} T7 items are expert-curated open-ended question--reference pairs; the released benchmark does not provide explicit evidence IDs for the target answer.}
    \end{tcolorbox}

\end{prettybox}


\endgroup

\clearpage
\section{Additional Experimental Results}
\label{app:extra_results}

This appendix collects the supplementary result views that were moved out of the main paper for space: the primary-backbone breakdowns used in the dataset-level analysis, and backbone-specific radar plots for the full evaluation set.

\begin{table}[t]
    \centering
    \small
    \setlength{\tabcolsep}{0.8pt}
    \renewcommand{\arraystretch}{1.25}

    \newcommand{\best}[1]{\textbf{#1}}
    \newcommand{\pos}[1]{\textcolor{teal}{+#1}}
    \newcommand{\negval}[1]{\textcolor{red}{#1}}
    \newcommand{\base}[1]{\textcolor{black}{#1}}
    \newcommand{\na}[0]{\textcolor{gray}{-}}
    \newcommand{\theader}{%
        \toprule
        \multirow{2}{*}{\textbf{System}} & \multicolumn{3}{c}{\textbf{Level I: Fact \& Entity}} & \multicolumn{2}{c}{\textbf{Level II: Temporal Logic}} & \multicolumn{2}{c}{\textbf{Level III: Insight}} \\
        \cmidrule(lr){2-4} \cmidrule(lr){5-6} \cmidrule(lr){7-8}
        & \textbf{T1 (Det.)} & \textbf{T2 (Ent.)} & \textbf{T3 (Time)} & \textbf{T4 (Rel.)} & \textbf{T5 (Con.)} & \textbf{T6 (Abs.)} & \textbf{T7 (Sum.)} \\
        \midrule
    }
    \newcommand{\groupheader}[1]{\multicolumn{8}{l}{\textit{#1}} \\
    }

    \begin{minipage}[t]{0.49\linewidth}
        \centering
        \textbf{(a) Overall (Primary Backbones)} \\ [2pt]
        \resizebox{\linewidth}{!}{%
            \begin{tabular}{lrrrrrrr}
                \theader
                \groupheader{$\bullet$ Backbone: Qwen3-32B}
                Base Model (Abs.) & \base{59.9} & \base{66.0} & \base{44.4} & \base{40.5} & \base{36.1} & \base{14.3} & \base{16.3} \\
                + Naive RAG       & \pos{8.8}  & \pos{4.8}  & \pos{3.9}  & \pos{1.8}  & \pos{2.7}  & \pos{2.8}  & \pos{1.7} \\
                + Mem0 (Entity)   & \best{\pos{10.5}} & \best{\pos{9.2}}  & \negval{-3.1} & \pos{1.9}  & \negval{-0.2} & \pos{2.6}  & \pos{1.2} \\
                + MemOS           & \pos{4.5}  & \pos{6.4}  & \best{\pos{8.3}}  & \best{\pos{6.6}}  & \best{\pos{5.0}}  & \best{\pos{3.9}}  & \best{\pos{3.0}} \\
                \midrule
                \groupheader{$\bullet$ Backbone: GPT-5-mini}
                Base Model (Abs.) & \base{65.4} & \base{71.5} & \base{54.1} & \base{47.3} & \base{42.3} & \base{18.6} & \base{19.6} \\
                + Naive RAG       & \pos{6.8}  & \pos{4.0}  & \pos{3.2}  & \pos{1.5}  & \pos{2.0}  & \pos{2.7}  & \pos{1.3} \\
                + Mem0 (Entity)   & \best{\pos{7.8}}  & \best{\pos{7.2}}  & \negval{-2.5} & \pos{1.4}  & \pos{0.3}  & \pos{2.1}  & \pos{1.2} \\
                + MemOS           & \pos{4.2}  & \pos{5.1}  & \best{\pos{6.7}}  & \best{\pos{5.6}}  & \best{\pos{5.2}}  & \best{\pos{2.9}}  & \best{\pos{3.0}} \\
                \bottomrule
            \end{tabular}%
        }
    \end{minipage}
    \hfill
    \begin{minipage}[t]{0.49\linewidth}
        \centering
        \textbf{(b) Dataset 1 (Flashbacks)} \\ [2pt]
        \resizebox{\linewidth}{!}{%
            \begin{tabular}{lrrrrrrr}
                \theader
                \groupheader{$\bullet$ Backbone: Qwen3-32B}
                Base Model (Abs.) & \base{58.1} & \base{64.2} & \base{38.5} & \base{35.6} & \base{34.3} & \base{13.8} & \base{15.1} \\
                + Naive RAG       & \pos{9.2}  & \pos{4.7}  & \pos{5.3}  & \pos{2.5}  & \pos{3.1}  & \best{\pos{6.2}}  & \pos{1.8} \\
                + Mem0 (Entity)   & \pos{10.5} & \pos{8.6}  & \negval{-3.5} & \pos{1.2}  & \negval{-0.2} & \pos{0.9}  & \pos{1.6} \\
                + MemOS           & \best{\pos{10.7}} & \best{\pos{9.7}}  & \best{\pos{10.4}} & \best{\pos{10.8}} & \best{\pos{4.2}} & \pos{3.4}  & \best{\pos{4.1}} \\
                \midrule
                \groupheader{$\bullet$ Backbone: GPT-5-mini}
                Base Model (Abs.) & \base{62.3} & \base{70.8} & \base{48.2} & \base{41.4} & \base{40.4} & \base{17.7} & \base{17.6} \\
                + Naive RAG       & \pos{7.5}  & \pos{3.8}  & \pos{4.5}  & \pos{1.8}  & \pos{2.4}  & \best{\pos{3.0}}  & \pos{0.9} \\
                + Mem0 (Entity)   & \best{\pos{8.6}}  & \best{\pos{6.7}}  & \negval{-1.7} & \pos{1.6}  & \negval{-0.4} & \pos{1.4}  & \pos{1.0} \\
                + MemOS           & \pos{8.4} & \pos{6.1}  & \best{\pos{8.2}} & \best{\pos{7.0}}  & \best{\pos{5.7}} & \pos{2.5}  & \best{\pos{4.4}} \\
                \bottomrule
            \end{tabular}%
        }
    \end{minipage}

    \vspace{0.8em}

    \begin{minipage}[t]{0.49\linewidth}
        \centering
        \textbf{(c) Dataset 2 (Event Dense)} \\ [2pt]
        \resizebox{\linewidth}{!}{%
            \begin{tabular}{lrrrrrrr}
                \theader
                \groupheader{$\bullet$ Backbone: Qwen3-32B}
                Base Model (Abs.) & \base{63.1} & \base{70.7} & \base{44.4} & \base{40.9} & \base{34.4} & \base{14.9} & \base{16.5} \\
                + Naive RAG       & \pos{8.8}  & \pos{5.7}  & \pos{1.9}  & \pos{1.1}  & \pos{2.0}  & \pos{0.8}  & \best{\pos{2.9}} \\
                + Mem0 (Entity)   & \best{\pos{9.3}}  & \best{\pos{8.9}}  & \negval{-4.1} & \pos{1.7}  & \negval{-1.5} & \pos{2.5}  & \negval{-0.8} \\
                + MemOS           & \pos{1.2}  & \pos{3.8}  & \best{\pos{7.3}}  & \best{\pos{5.7}}  & \best{\pos{5.3}}  & \best{\pos{3.4}}  & \best{\pos{3.0}} \\
                \midrule
                \groupheader{$\bullet$ Backbone: GPT-5-mini}
                Base Model (Abs.) & \base{68.5} & \base{73.0} & \base{55.4} & \base{49.8} & \base{42.5} & \base{19.6} & \base{20.3} \\
                + Naive RAG       & \pos{5.7}  & \pos{4.0}  & \pos{0.8}  & \pos{2.6}  & \negval{-0.1} & \pos{1.8}  & \pos{1.0} \\
                + Mem0 (Entity)   & \best{\pos{7.1}}  & \best{\pos{6.2}}  & \negval{-3.2} & \pos{1.8}  & \negval{-2.1} & \pos{1.5}  & \pos{0.2} \\
                + MemOS           & \pos{2.5}  & \pos{4.2}  & \best{\pos{5.6}}  & \best{\pos{5.6}}  & \best{\pos{4.0}}  & \best{\pos{3.0}}  & \best{\pos{3.7}} \\
                \bottomrule
            \end{tabular}%
        }
    \end{minipage}
    \hfill
    \begin{minipage}[t]{0.49\linewidth}
        \centering
        \textbf{(d) Dataset 3 (Mind)} \\ [2pt]
        \resizebox{\linewidth}{!}{%
            \begin{tabular}{lrrrrrrr}
                \theader
                \groupheader{$\bullet$ Backbone: Qwen3-32B}
                Base Model (Abs.) & \base{58.6} & \base{63.2} & \base{50.4} & \base{45.1} & \base{39.5} & \base{14.2} & \base{17.3} \\
                + Naive RAG       & \pos{8.8}  & \pos{3.7}  & \pos{4.4}  & \pos{1.6}  & \pos{3.0}  & \pos{1.2}  & \pos{0.4} \\
                + Mem0 (Entity)   & \best{\pos{11.5}} & \best{\pos{9.8}}  & \negval{-0.9} & \pos{2.7}  & \pos{1.3}  & \pos{4.3}  & \best{\pos{2.7}} \\
                + MemOS           & \pos{1.5}  & \pos{6.6}  & \best{\pos{7.0}}  & \best{\pos{3.3}}  & \best{\pos{5.6}}  & \best{\pos{4.9}}  & \pos{1.8} \\
                \midrule
                \groupheader{$\bullet$ Backbone: GPT-5-mini}
                Base Model (Abs.) & \base{65.3} & \base{71.1} & \base{58.6} & \base{51.2} & \base{44.1} & \base{18.5} & \base{20.9} \\
                + Naive RAG       & \pos{7.2}  & \pos{3.9}  & \pos{3.0}  & \negval{-0.4} & \pos{3.5}  & \pos{2.4}  & \pos{0.7} \\
                + Mem0 (Entity)   & \best{\pos{7.8}}  & \best{\pos{8.3}}  & \negval{-2.5} & \pos{0.2}  & \pos{0.1}  & \best{\pos{3.5}}  & \best{\pos{1.4}} \\
                + MemOS           & \pos{1.8}  & \pos{5.3}  & \best{\pos{2.1}}  & \best{\pos{3.6}}  & \best{\pos{4.7}}  & \pos{1.6}  & \pos{0.9} \\
                \bottomrule
            \end{tabular}%
        }
    \end{minipage}

    \caption{Per-dataset breakdowns for the two primary backbones. Panel~(a) reports the overall comparison for Qwen3-32B and GPT-5-mini; panels~(b)--(d) provide the dataset-level views referenced in the main-text analysis. Base rows show absolute scores, while the memory variants are reported as deltas relative to the corresponding base model.}
    \label{tab:full_results}
\end{table}

\begin{figure}[t]
    \centering
    \includegraphics[width=0.95\linewidth]{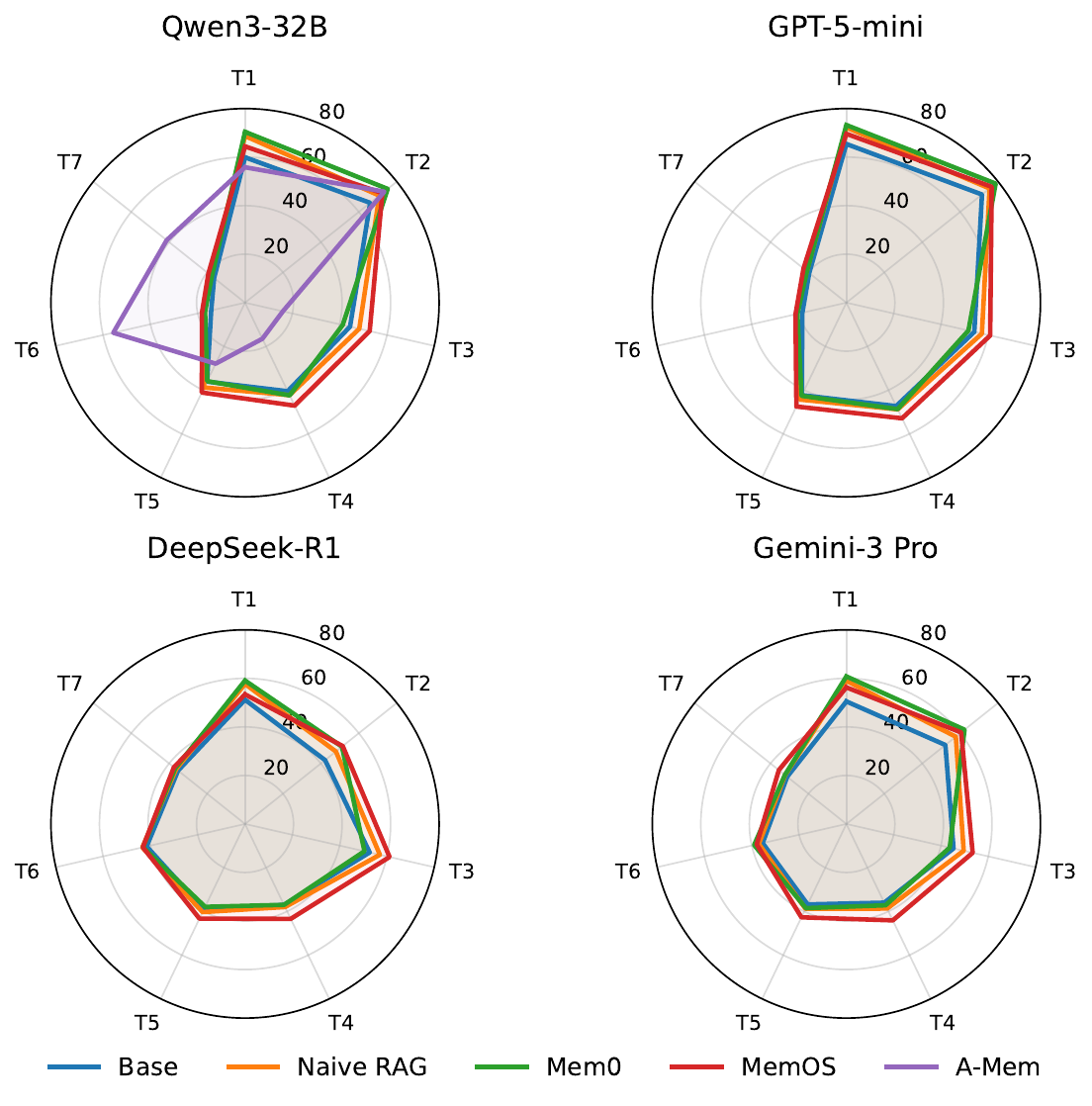}
    \caption{Backbone-specific radar plots for the overall evaluation results. All four backbone families include Base, Naive RAG, Mem0, and MemOS; Qwen3-32B additionally includes A-Mem. The figure makes the same pattern visible as Table~\ref{tab:main_results}: retrieval-heavy variants help T1--T2, while MemOS yields broader gains on T3--T5.}
    \label{fig:radar_charts}
\end{figure}

\clearpage
\section{Human Evaluation Details}
\label{app:human_eval_details}

\begin{table}[h]
    \centering
    \small
    \setlength{\tabcolsep}{7pt}
    \renewcommand{\arraystretch}{1.05}
    \begin{tabular}{lcc}
        \toprule
        \textbf{Level} & \textbf{Human} & \textbf{Best Model} \\
        \midrule
        I: Fact \& Memory & 96.5 & 75.4 \\
        II: Logic \& Causality & 88.0 & 62.5 \\
        III: Insight & 83.5 & 22.6 \\
        \bottomrule
    \end{tabular}
    \caption{Human versus best-model performance aggregated by evaluation level. Human answers are graded by the same blinded LLM-as-a-Judge pipeline used for model outputs.}
    \label{tab:human_eval}
\end{table}

\paragraph{Annotator demographics and ethics.}
The human study used three expert annotators with M.A.-level training in literature or linguistics. All annotators were compensated at a reasonable rate consistent with standard expert-annotation practice. The study was designed as an expert reading and evidence-grounded question-answering task rather than open-ended crowd annotation.

\paragraph{Evaluation materials and interface.}
For each item, annotators were shown the reconstructed cognitive stream used by the benchmark. To preserve global narrative context and avoid penalizing humans for missing book-level structure, they could also consult the original novel text at any time during evaluation. This setup ensured that annotators had access both to the benchmark representation and to the full narrative evidence from which it was derived.

\paragraph{Instructions and scoring pipeline.}
Annotators were instructed to answer each question only with claims supported by textual evidence. They did not score one another's outputs. Instead, their written answers were passed through the same level-specific LLM-as-a-Judge pipeline used for model responses so that human and model performance were evaluated under an identical rubric. During scoring, the judge was blind to whether an answer came from a human annotator or from a model system.

\paragraph{Agreement interpretation.}
Under this protocol, the relevant reliability quantity is the alignment between the rubric-based LLM judge and expert consensus on the subjective subset, rather than pairwise agreement among humans grading one another. The reported $\kappa > 0.75$ therefore reflects strong alignment between the judge's scoring behavior and expert consensus, which is why we do not additionally report a separate table of pairwise human--human Cohen's Kappa values.

\normalsize
\end{document}